\documentclass[a4paper,fleqn]{cas-dc}

\usepackage{graphicx} 
\usepackage{textcomp,makecell} 
\usepackage{xcolor,comment} 
\usepackage[T1]{fontenc} 
\usepackage[utf8]{inputenc} 
\usepackage{algpseudocode} 
\usepackage{float} 
\usepackage{soul} 
\usepackage[normalem]{ulem} 
\usepackage[ruled,vlined,linesnumbered]{algorithm2e} 
\usepackage{subcaption} 
\usepackage[authoryear,longnamesfirst]{natbib} 
\usepackage{orcidlink}

\begin{document}
\let\WriteBookmarks\relax

\shorttitle{Robust Reconstruction of Power Lines from LiDAR Data}
\shortauthors{Gaha et al.}

\title[mode=title]{Robust Reconstruction of Power Lines from Incomplete Mobile and Aerial LiDAR Data}

\author[1]{Mohamed Gaha\ \orcidlink{0000-0002-4574-4623}}
\ead{gaha.mohamed@hydroquebec.com}

\author[1,2]{Damos Ayobo Abongo\ \orcidlink{0009-0006-9666-5665}}
\ead{ayoboabongo.damos@hydroquebec.com}

\author[1]{Philippe Massicotte\ \orcidlink{0000-0002-7170-2670}}
\ead{massicotte.philippe@hydroquebec.com}

\author[1,2]{Eya Ben Kacem}
\ead{benkacem.eya@hydroquebec.com}

\author[2]{Wael Jaafar\ \orcidlink{0000-0003-4378-9999}}
\cormark[1]
\ead{wael.jaafar@etsmtl.ca}

\author[2,3]{Rami Langar\ \orcidlink{0000-0002-4674-6700}}
\ead{rami.langar@etsmtl.ca}

\author[1]{Christian Buteau}
\ead{buteau.christian@hydroquebec.com}

\affiliation[1]{organization={Hydro-Quebec},
  city={Varennes},
  postcode={J3X 1S1},
  state={Quebec},
  country={Canada}}

\affiliation[2]{organization={Software and IT Engineering Department, \'{E}cole de Technologie Sup\'{e}rieure},
  city={Montreal},
  postcode={H3C 1K3},
  state={Quebec},
  country={Canada}}

\affiliation[3]{organization={LIGM-CNRS UMR 8049, University Gustave Eiffel},
  city={Marne-la-Vall\'{e}e},
  postcode={F-77420},
  country={France}}

\cortext[1]{Corresponding author}
% Allow figure-only pages without requiring them to be almost full
\renewcommand{\floatpagefraction}{0.5}
\renewcommand{\dblfloatpagefraction}{0.5}
% Allow more space for figures on pages containing text
\renewcommand{\topfraction}{0.95}
\renewcommand{\bottomfraction}{0.95}
\renewcommand{\textfraction}{0.05}
\renewcommand{\dbltopfraction}{0.95}

\begin{abstract}
Accurate vegetation-encroachment assessment and effective risk
mitigation require electric utilities to precisely locate and characterize distribution-network assets, including poles and power lines, and to distinguish among line types such as medium-voltage, low-voltage, neutral, and telecommunication lines. Mobile and aerial Light Detection and Ranging (LiDAR) point clouds provide high-resolution 3D representations of these assets and their surroundings, including vegetation, buildings, and roads. However, detecting and reconstructing power lines directly from LiDAR data remains challenging since conductor returns are often sparse, incomplete, noisy, or occluded by vegetation. These challenges are
pronounced in distribution networks, where relatively low pole heights and complex line configurations place conductors in close proximity to trees. Moreover, mobile and aerial LiDAR
point clouds differ substantially in point density, viewing geometry,
spatial coverage, and occlusion patterns. In this work, we propose a robust and automated reconstruction method that combines scaled density-based clustering, robust polynomial
fitting, and iterative fragment merging to recover individual conductors under severe data incompleteness and occlusion. Following semantic classification, the proposed reconstruction pipeline processes mobile and aerial LiDAR data using the same implementation and parameter configuration, without requiring
modality-specific reconstruction code or parameter tuning. We evaluate our method through controlled degradation experiments,
comparative benchmarking, parameter sensitivity analysis, and
large-scale validation on mobile and aerial LiDAR datasets. On the aerial dataset, the proposed method achieves a precision of 96.87\%, a recall of 96.66\%, an F$_1$-score of 96.76\%, and a geometric root mean square error (RMSE) of 0.127~m. On the mobile dataset, for which independent conductor-level reference annotations were unavailable, the method achieves a point-to-model RMSE of
0.045~m and a reconstruction coverage of 95.27\%. To support reproducibility and facilitate further research, we publicly release the source code, documentation, and evaluation data available on \url{https://github.com/varen-project/varen}.
\end{abstract}


\maketitle

\section{Introduction}
\label{sec:introduction}

Reconstructing power-line geometry is essential for electric
utilities to assess vegetation encroachment and mitigate risks to
distribution infrastructure. Light Detection and Ranging (LiDAR) point
clouds provide high-resolution 3D representations of utility assets and their surroundings, including vegetation, buildings, and roads~\cite{Waykar2022}. Therefore, these data offer significant potential for automated power-network inspection, monitoring, and asset management. Nevertheless, reliable power-line detection and reconstruction from LiDAR point clouds remain challenging, particularly in distribution networks. Conductors and poles are often located close to trees and branches, resulting in frequent occlusions, sparse conductor returns, and confusion between power-line points and nearby vegetation.

Traditional reconstruction methods, including voxel-based approaches~\cite{Jwa2009}, often exhibit limited robustness to
incomplete or noisy point clouds and may incur substantial computational
costs on large datasets. More recent context-aware approaches~\cite{2025IJSTA,gaha2021,Guo2016,Shi2020} exploit pole locations, span geometry, and mathematical conductor models to estimate power-line shapes. Although they represent important advances, their performance may deteriorate under severe occlusion or substantial data incompleteness. Recent reviews have shown that many existing power-line extraction and reconstruction methods were evaluated on relatively small datasets and that comparatively few studies addressed complex multi-conductor configurations \cite{Munir2023PowerLineReview}. Furthermore, most have been evaluated on relatively limited datasets, leaving their robustness and scalability for large-scale power-network reconstruction insufficiently investigated. 

Existing methods are designed for a single LiDAR acquisition modality. Mobile LiDAR provides dense and detailed ground-level measurements, but its observations may be affected by
roadside obstacles, uneven sampling, and limited visibility of elevated
conductors. In contrast, aerial LiDAR provides broader spatial coverage
from a predominantly top-down perspective, but conductor returns may be sparser and more difficult to distinguish from nearby vegetation. These differences in point density, viewing geometry, spatial coverage, noise, and occlusion patterns complicate the development of a reconstruction method that performs consistently across both modalities. Consequently, a method developed for one modality may require substantial parameter retuning or implementation changes before being applied to the other.

Recent learning-based studies have investigated power-line point extraction from both aerial and mobile LiDAR data \cite{Kumar2024DeepLearningPowerlines}, unified conductor-level geometric reconstruction across these acquisition modalities remains insufficiently investigated. To the best of our knowledge, no previous power-line reconstruction
method has been specifically designed and extensively evaluated on both
mobile and aerial LiDAR data using the same reconstruction implementation and parameter configuration. Addressing this limitation is important for utilities that acquire heterogeneous LiDAR datasets and require a scalable solution without modality-specific reconstruction code or extensive manual calibration. Hence, we propose a robust and automated reconstruction framework that combines adaptive coordinate
prescaling, density-based clustering, robust polynomial fitting, and iterative fragment merging to recover individual conductors under severe
occlusion and data incompleteness. The main contributions of this work
are as follows:
\begin{enumerate}
    \item We develop a unified reconstruction framework for mobile and aerial LiDAR point clouds. Following modality-specific semantic classification, the framework uses the same reconstruction
    implementation and parameter configuration for both acquisition
    modalities.

    \item We introduce a robust reconstruction method that combines
    adaptive coordinate prescaling, HDBSCAN clustering, Huber-weighted polynomial fitting, iterative fragment merging, and a rescue stage for rejected conductor candidates.

    \item We conduct an extensive evaluation that includes controlled degradation experiments, comparisons with existing reconstruction pipelines, and a comprehensive parameter-sensitivity analysis. The large-scale evaluation covers 1,620 mobile LiDAR spans over 68~km and 10,870 aerial LiDAR spans over 382~km.

    \item We publicly release the source code, documentation, and
    evaluation data. To the best of our knowledge, these resources
    constitute the first publicly available benchmark that includes
    power-line reconstruction data from both mobile and aerial LiDAR on \url{https://github.com/varen-project}.
    %This release supports reproducibility, facilitates comparative evaluation, and encourages further research.
\end{enumerate}

The remainder of this paper is organized as follows.
Section~\ref{sec:related-work} reviews related work on LiDAR-based
power-line detection and reconstruction. Section~\ref{sec:methodology}
presents the proposed reconstruction method.
Section~\ref{sec:results} describes the controlled robustness
experiments, large-scale evaluations, comparative benchmarking, and
parameter sensitivity analysis. Finally, Section~\ref{sec:conclusion}
summarizes the main findings, limitations, and directions for future
research.

\section{Related Work}
\label{sec:related-work}

This section reviews representative methods for detecting, extracting,
and reconstructing power lines from LiDAR point clouds. Particular
attention is given to their assumptions, methodological strengths, and limitations regarding vegetation-induced occlusions, incomplete
conductor observations, closely spaced phases, and differences between
LiDAR acquisition modalities.

Jwa et al.~\cite{Jwa2009} introduced the Voxel-based Piece-wise Line
Detector (VPLD) for the automatic 3D reconstruction of power lines from airborne laser scanning (ALS) LiDAR data. Their method partitioned the point cloud into sub-cubic voxels and grouped neighboring voxels likely to contain conductor points. The resulting components were connected, and the parameters of a catenary model were iteratively refined to produce continuous conductor reconstructions. This piecewise strategy recovered
conductor geometry from locally detected observations, but its
performance deteriorated when conductor returns were sparse or heavily occluded by vegetation. Its voxel-based iterative processing could also incur substantial computational costs on large datasets.

Liang and Zhang~\cite{Liang2011} proposed a method for extracting power lines from ALS LiDAR data. Their method identified candidate conductor points through geometric filtering and applied a Hough-like line detector to organize and connect the extracted observations. The resulting points were modeled in 3D space using a
parabolic equation whose parameters were estimated by least-squares
fitting. However, the method became less effective under substantial
occlusion because large observation gaps prevented fragments belonging
to the same conductor from being correctly associated. 

Guo et al.~\cite{Guo2016} developed a contextual reconstruction method
for ALS data. Their approach used the
locations of two pylons delimiting a span to constrain the conductor
search region. This region was partitioned into equally spaced cuboids, and candidate wire points were grouped according to their geometric characteristics and a neighborhood-based ratio. The conductors were then reconstructed using quadratic fitting and an iterative RANSAC
procedure~\cite{Fischler}. By incorporating span-level contextual
information, their method reconstructed conductors even when portions of their LiDAR observations were missing. Nevertheless, it assumed that the conductor configuration remained sufficiently consistent throughout each span, which limited its applicability to spans containing
substantial horizontal or vertical configuration changes.

For mobile laser scanning (MLS) data, Gaha et al.~\cite{gaha2021} proposed a contextual method for detecting and
classifying utility poles and power lines. The search for conductor
candidates was restricted to points located near a plane defined by two
consecutive poles, thereby reducing the number of irrelevant observations and the processing complexity. RANSAC was used to extract approximately linear conductor segments, while RGB information was used to filter the resulting candidates. Subsequently, each conductor was modeled using three points, i.e., two points associated with consecutive poles and one point selected from the remaining candidates. The method was
evaluated on 98 poles and achieved a recall of 98\% and an accuracy of
84\%. However, the experiments were limited to single-phase configurations, and the three-point model was sensitive to candidate-point misclassification, particularly under severe occlusion. In a subsequent study, Massicotte et al.~\cite{Massicotte2024} extended
this approach by combining mobile LiDAR and image information to
support utility-asset inventory. The reported accuracy ranged from
85\% to 95\%. 

Also, Shi et al.~\cite{Shi2020} investigated power-line extraction from MLS data in urban environments. Their method combined geometric feature extraction and clustering to identify power-line points and support hazard analysis. Although designed for dense ground-level observations, the method was not evaluated on aerial LiDAR data. 

More recently, Cao et al.~\cite{2025IJSTA} proposed an automated method for extracting power lines from ALS point clouds using connected-component segmentation and contextual geometric information. Their approach exploited the alignment of linear features with the span direction to distinguish conductor points from unrelated structures. To recover fragmented conductor segments, the authors introduced a graph-cut energy-minimization algorithm and a slice-based segmentation procedure that separated power lines from nearby pylons, insulators, and jumper wires. The method was evaluated on nine datasets covering more than 80~km of transmission corridors with voltage classes ranging from 220 to 800~kV. It achieved an average precision of 99.77\%, an average recall of 99.81\%, and an average F$_1$-score of 99.79\%. Despite its high reported performance, the method of Cao et
al.~\cite{2025IJSTA} relied on distance-based connectivity assumptions
that are unreliable when gaps within a conductor are larger than
the separation between neighboring conductors. This condition commonly
occurs in distribution networks, where phases are closely spaced, and
conductor observations may be fragmented by vegetation. Although these connectivity assumptions were suitable for transmission corridors with widely separated phases, they are less appropriate for dense
distribution-line configurations.

Maset and Fusiello \cite{Maset2024Vectorization} proposed a complete airborne-LiDAR processing pipeline combining deep-learning-based semantic segmentation, DBSCAN-based pylon identification, and J-Linkage multi-model fitting for conductor vectorization. The method achieved an F$_1$-score of 98.5\% for vectorized cables on a 36-km transmission-line dataset. However, it was evaluated on airborne high-voltage infrastructure and required explicit catenary-based multi-model fitting.

The above studies demonstrated the value of combining contextual information with geometric conductor modeling. However, most methods are sensitive to severe occlusions, fragmented observations, or closely spaced conductors. When neighboring conductors are closely spaced, the extraction of individual conductors becomes challenging \cite{Shen2024PowerlineReview}. Generally, they were designed and evaluated for a single LiDAR acquisition modality. 
Consequently, their robustness and scalability across mobile and aerial
LiDAR data remain insufficiently investigated. Table~\ref{tab:powerline-lidar-review} summarizes the objectives, principal techniques, data sources, reported performance, and main
limitations of the reviewed methods. The reported values should be interpreted with caution because the studies used different datasets,
voltage classes, evaluation units, and metric definitions. 
%Accordingly, these values do not constitute a direct comparative benchmark.

\begin{table*}[pos=!htb]
\centering
\caption{Comparison of LiDAR-based power-line detection and
reconstruction methods. 
%Reported values are taken from the corresponding studies and are not directly comparable because of differences in datasets, evaluation protocols, and metric definitions.
}
\label{tab:powerline-lidar-review}
\small
\setlength{\tabcolsep}{3.5pt}
\renewcommand{\arraystretch}{1.15}

\begin{tabular}{
    p{1.7cm}
    p{1.8cm}
    p{2.7cm}
    p{3.2cm}
    p{1.0cm}
    p{2.1cm}
    p{3.2cm}
}
\hline
\textbf{Reference} &
\textbf{Objective} &
\textbf{Main technique} &
\textbf{Context or model} &
\textbf{Data} &
\textbf{Reported performance} &
\textbf{Main limitation} \\
\hline

%Jwa et al.
~\cite{Jwa2009} &
3D power-line reconstruction &
Voxel-based piecewise detection and catenary fitting &
Local voxel grouping and iterative catenary refinement &
ALS &
Accuracy: 93.8\% &
Sensitive to vegetation-induced occlusion and potentially
computationally expensive \\
\hline

%Liang and Zhang
~\cite{Liang2011} &
Power-line extraction and modeling &
Geometric filtering, Hough-like detection, and least-squares fitting &
Three-dimensional parabolic conductor model &
ALS &
Accuracy: 96.5\% &
Performance deteriorated when conductor observations contained large
gaps \\
\hline

%Guo et al.
~\cite{Guo2016} &
Robust conductor reconstruction &
Cuboid segmentation, quadratic fitting, and iterative RANSAC &
Span constrained by two pylons &
ALS &
Accuracy: 94.9\% &
Assumed a sufficiently consistent conductor configuration within each
span \\
\hline

%Gaha et al.
~\cite{gaha2021} &
Pole and power-line detection &
RANSAC-based segment extraction and RGB filtering &
Plane defined by consecutive poles and three-point modeling &
MLS &
Recall: 98\%; accuracy: 84\% &
Limited to single-phase configurations and sensitive to candidate-point
errors \\
\hline

%Shi et al.
~\cite{Shi2020} &
Urban power-line extraction and hazard analysis &
Geometric feature extraction and clustering &
Urban roadside distribution-network context &
MLS &
Accuracy: 91\% &
Designed for mobile LiDAR and not evaluated on aerial data \\
\hline

%Massicotte et al.
~\cite{Massicotte2024} &
Utility-asset inventory &
Combined image and LiDAR processing &
Contextual detection of utility assets &
MLS &
Accuracy: 85--95\% &
Limited implementation details restricted reproducibility and analysis \\
\hline

%Cao et al.
~\cite{2025IJSTA} &
Automated extraction in transmission corridors &
Connected components, graph cuts, slice-based segmentation, and energy
minimization &
Span alignment and contextual geometric information &
ALS &
Precision: 99.77\%; recall: 99.81\%; F$_1$: 99.79\% &
Distance-based connectivity may fail for closely spaced distribution
conductors \\
\hline
%Maset and Fusiello
~\cite{Maset2024Vectorization} &
Automatic power-line vectorization &
FCN segmentation, DBSCAN and J-Linkage multi-model fitting &
contextual geometric information &
ALS &
F$_1$: 98.5\% &
Performance on closely spaced distribution conductors was not assessed \\
\hline

\textbf{Proposed method} &
Power-line detection and reconstruction &
Adaptive prescaling, HDBSCAN, robust polynomial fitting, and iterative
fragment merging &
Unified reconstruction of mobile and aerial LiDAR spans &
\makecell{ALS\\MLS} &
ALS: precision 96.87\%; recall 96.66\%; F$_1$: 96.76\% &
Independent conductor annotations were unavailable for the mobile
LiDAR evaluation \\
\hline

\end{tabular}
\end{table*}

\section{Proposed Power-Line Detection and Reconstruction Method}
\label{sec:methodology}

The complete workflow consists of a modality-specific semantic
classification stage followed by a unified four-stage reconstruction
pipeline: 1) scaled clustering and geometric filtering, 2) robust
parametric conductor modeling, 3) iterative fragment merging and
conductor acceptance, and 4) rescue of rejected fragments. The semantic
classification procedure differs between mobile and aerial LiDAR data,
whereas all subsequent reconstruction stages use the same implementation and parameter configuration.

\subsection{Semantic Classification of Poles and Power Lines}
\label{sec:semantic-classification}

LiDAR semantic classification remains challenging because of variations in point density and intensity, occlusions, measurement noise, and scanning artifacts. Learning-based methods can address some of these limitations by jointly exploiting geometric, contextual, and RGB information.

\begin{figure}[pos=!htb]
\centering
\includegraphics[width=\columnwidth]{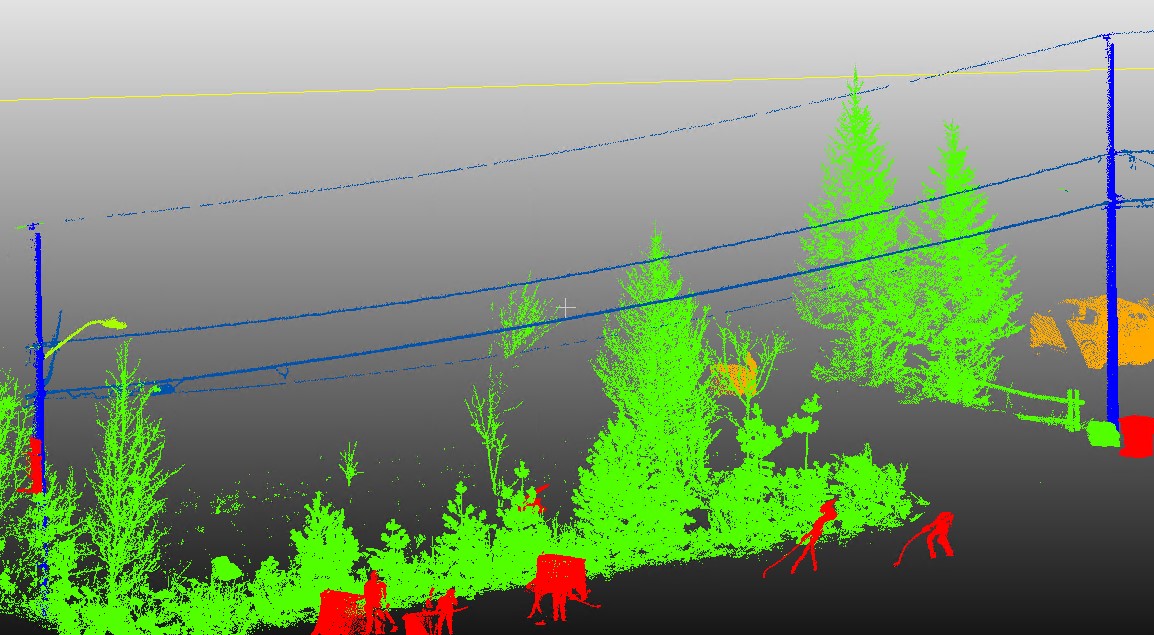}
\caption{Example of a mobile LiDAR span classified using U-Next.
Vegetation is shown in green, power lines in light blue, poles in dark
blue, vehicles in orange, public-lighting assets in light green, and
other objects in red.}
\label{fig:classified-lidar}
\end{figure}

For mobile LiDAR data, we trained a U-Next semantic-segmentation model~\cite{ZENG2025104309} on an internally annotated dataset. A small
subset of point-cloud tiles was first manually annotated and used to train an initial model. The model was then applied to additional tiles, and its preliminary predictions were reviewed and corrected by a human annotator. The corrected annotations were incorporated into subsequent
training iterations. This model-assisted annotation procedure was
repeated until the classification performance stabilized. The final
model achieved an average Intersection over Union (mIoU) of 87\%
over the critical power-line, vegetation, and pole classes. An example of the resulting semantic classification is shown in
Fig.~\ref{fig:classified-lidar}.

The trained model classified mobile LiDAR points into seven categories:
vegetation, wires, poles, transformers, buildings, vehicles, and other objects. The annotated dataset contained approximately 154.5 million
points collected in rural and semi-urban environments. As shown in
Table~\ref{tab:mobile-class-distribution}, the dataset was highly
imbalanced. Indeed, transformers and poles represented approximately 0.05\%
and 1.23\% of all annotated points, respectively. The data were divided
geographically into mutually exclusive training, validation, and test regions using a 60\%/15\%/25\% split. This geographic separation prevented points from the same span or acquisition segment from appearing in multiple subsets.

\begin{table}[pos=!htb]
\centering
\caption{Class distribution in the HQ-2025 mobile LiDAR dataset. Percentages indicate the proportion of points within each split.}
\label{tab:mobile-class-distribution}
\small
\setlength{\tabcolsep}{3.5pt}
\renewcommand{\arraystretch}{1.1}
\begin{tabular}{lcccc}
\hline
\textbf{Class} &
\textbf{Train (\%)} &
\textbf{Val. (\%)} &
\textbf{Test (\%)} &
\textbf{Count (M)} \\
\hline
Vegetation  & 89.79 & 72.54 & 89.98 & 134.6 \\
Wire        & 1.31  & 3.24  & 1.54  & 2.6   \\
Pole        & 0.98  & 2.25  & 1.16  & 1.9   \\
Transformer & 0.04  & 0.11  & 0.03  & 0.07  \\
Building    & 5.71  & 17.29 & 6.26  & 11.9  \\
Vehicle     & 1.38  & 1.69  & 0.83  & 2.0   \\
Other       & 0.79  & 2.89  & 0.20  & 1.5   \\
\hline
\textbf{Total points} &
\textbf{91.0 M} &
\textbf{24.4 M} &
\textbf{39.1 M} &
\textbf{154.5 M} \\
\hline
\end{tabular}
\end{table}

For aerial LiDAR data, we used either provider-supplied semantic labels
or labels generated using commercial software such as Global
Mapper~\cite{global_mapper_pro}. Developing a new aerial semantic
classification model was outside the scope of this work, which focused
on conductor reconstruction after candidate wire points had been
identified. Hence, the semantic preprocessing stage was modality-specific, i.e., mobile LiDAR data were classified using U-Next, whereas aerial LiDAR data used provider-supplied or commercially generated labels. Following this preprocessing stage, the same reconstruction implementation and parameter configuration were applied
to both modalities.

\subsection{Scaled Clustering and Geometric Filtering}
\label{sec:scaled-clustering}

Semantic segmentation assigns all wire points within a span to the same class without distinguishing individual conductors, phases, voltage levels, or telecommunication cables. To separate these observations, we applied Hierarchical Density-Based Spatial Clustering of Applications with Noise (HDBSCAN)~\cite{McInnes2017HDBSCAN}. The latter was selected instead of the original DBSCAN algorithm \cite{Ester1996DBSCAN} because it can identify clusters with varying densities while labeling isolated observations as noise. However, direct clustering in the original coordinate system remained difficult because a conductor could extend over several tens of meters while adjacent conductors
could be separated by less than one meter. Consequently, we transformed the
points into a scaled, span-aligned coordinate system before clustering. 
Let $\mathbf{p}_i =\begin{bmatrix}
x_i & y_i & z_i \end{bmatrix}^{\top}$ denote the \(i\)-th candidate wire point, and $\mathbf{p}_i^{xy} =
\begin{bmatrix} x_i & y_i \end{bmatrix}^{\top}$ its horizontal coordinates. Principal Component Analysis (PCA) was applied to the horizontal point coordinates. The first principal direction, denoted by \(\mathbf{u}\in\mathbb{R}^{2}\), approximated the along-span direction. A perpendicular unit vector \(\mathbf{n}\in\mathbb{R}^{2}\) defined the cross-span direction as follows: 
\begin{equation}
\mathbf{u}^{\top}\mathbf{n}=0,
\qquad
\lVert\mathbf{u}\rVert_2
=
\lVert\mathbf{n}\rVert_2
=
1.
\end{equation}
Using the horizontal centroid
\(\bar{\mathbf{p}}^{xy}\) and mean elevation \(\bar{z}\), each point
was projected into the local coordinate system as
\begin{align}
s_i &=
\left(
\mathbf{p}_i^{xy}-\bar{\mathbf{p}}^{xy}
\right)^{\top}\mathbf{u},
\label{eq:along-span-coordinate}\\
n_i &=
\left(
\mathbf{p}_i^{xy}-\bar{\mathbf{p}}^{xy}
\right)^{\top}\mathbf{n},
\label{eq:cross-span-coordinate}\\
h_i &= z_i-\bar{z},
\label{eq:centered-height}
\end{align}
where \(s_i\), \(n_i\), and \(h_i\) represented the along-span,
cross-span, and centered vertical coordinates, respectively. The
transformed point used for clustering is \begin{equation}
\widetilde{\mathbf{p}}_i =
\begin{bmatrix}
w_s s_i &
w_n n_i &
w_z h_i
\end{bmatrix}^{\top},
\label{eq:scaled-point}
\end{equation}
where \(w_s\), \(w_n\), and \(w_z\) controlled the relative influence
of the three coordinates. We used \(w_n=2\) and \(w_z=1\) to amplify lateral
separations between neighboring conductors while preserving vertical separations. For example,
setting \(w_n=2\) caused a lateral separation of 1~m to contribute
2~m to the scaled distance. In contrast, \(w_s\) compressed distances
along the span direction, thus facilitating the association of
fragments belonging to the same conductor. Hence, the transformation 
reduced along-span gaps while increasing the separation between
adjacent phases. %Fig.~\ref{fig:normalized-scaled-data} illustrates the effect of this transformation.

%\begin{figure*}[pos=!htb]
%\centering
%\includegraphics[width=0.8\linewidth]{imgs/Original Data.jpg}
%\vspace{1mm}
%\includegraphics[width=0.8\linewidth]{imgs/Scaled Normalized Data.jpg}
%\caption{ \textcolor{red}{Three-phase conductor observations before (top) and after (bottom) span-aligned prescaling. Compressing the along-span coordinate and expanding the cross-span and vertical coordinates improved conductor
%separation and increased the scaled distance between isolated points
%and valid conductor observations.}}
%\label{fig:normalized-scaled-data}
%\end{figure*}

The along-span weight \(w_s\) was adapted to the span length and the
HDBSCAN \texttt{cluster\_selection\_epsilon} parameter. Let \(L\)
denote the span length, \(\epsilon_{\mathrm{HDBSCAN}}\) the HDBSCAN
cluster-selection epsilon, and \(\rho_s=20\) a fixed scaling coefficient. The target scaled length is defined as
\begin{equation}
\ell_{\mathrm{target}}
=
\rho_s \times \epsilon_{\mathrm{HDBSCAN}},
\label{eq:target-scaled-length}
\end{equation}
and the preliminary along-span weight is 
\begin{equation}
w_s^{\star}
=
\frac{\ell_{\mathrm{target}}}{L}.
\label{eq:preliminary-along-weight}
\end{equation}
The final weight was obtained by clipping this value within predefined
bounds:
\begin{equation}
w_s
=
\operatorname{clip}
\left(
w_s^{\star},
w_{s,\min},
w_{s,\max}
\right),
\label{eq:adaptive-along-weight}
\end{equation}
where \(w_{s,\min}=0.05\) and \(w_{s,\max}=1.0\). Thus, the along-span
coordinate was scaled by between 5\% and 100\% of its original value,
with stronger compression applied to longer spans.

HDBSCAN was applied with \(\epsilon_{\mathrm{HDBSCAN}}=0.3\), a minimum number of samples \(k_{\min}=3\).

\begin{figure*}[pos=!htb]
\centering
\includegraphics[width=0.9\linewidth]
{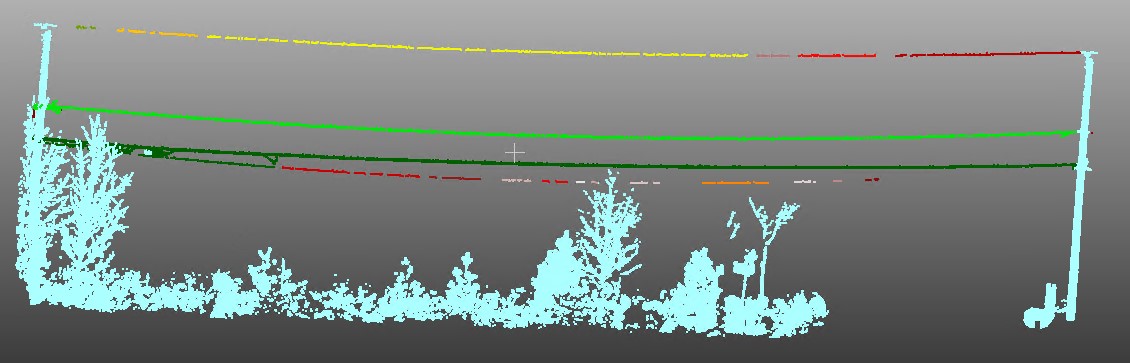}
\caption{Conductor fragments obtained using span-aligned prescaling,
HDBSCAN clustering, and geometric filtering. Each color represents a
different cluster.}
\label{fig:clustered-span}
\end{figure*}

After HDBSCAN clustering using the implementation of McInnes et al. \cite{McInnes2017HDBSCAN}, geometric and contextual criteria are used to remove likely classification errors. Clusters having a horizontal extent below 0.1~m are rejected. Also, PCA is applied to each cluster to assess its dominant orientation. Clusters whose principal direction is inconsistent with the span direction (absolute cosine below 0.85, i.e., a deviation larger than approximately \(32^{\circ}\)) are considered unlikely to represent conductors and are removed. This filtering reduces vegetation points incorrectly assigned to the wire class. Fig.~\ref{fig:clustered-span} shows the remaining clusters for the example in Fig.~\ref{fig:classified-lidar}.

\subsection{Robust Parametric Conductor Modeling}
\label{sec:parametric-model}
After clustering, observations belonging to a physical conductor can
remain divided among several spatially separated fragments, especially
when vegetation occluded parts of the conductor or when the LiDAR data
are sparse. A conductor model is therefore fitted to each cluster to identify geometrically compatible fragments. A catenary model can be fitted directly to the observed point coordinates without explicitly specifying the conductor’s physical properties such as conductor tension, temperature, and linear mass. We adopt a quadratic approximation for computational simplicity and stable fitting of highly fragmented observations. This choice is further supported by our experiments, in which quadratic fitting yielded slightly better reconstruction performance than catenary fitting. Accordingly, each conductor is represented by a span-aligned parametric model whose lateral and vertical components are modeled independently as linear and quadratic functions, respectively, of the local along-span coordinate \(s\) as follows:
\begin{align}
\widehat{n}(s)
&=
\beta_1s+\beta_0,
\label{eq:lateral-model}\\
\widehat{h}(s)
&=
\gamma_2s^2+\gamma_1s+\gamma_0.
\label{eq:vertical-model}
\end{align}
The corresponding point in the original coordinate system is
\begin{equation}
\widehat{\mathbf{p}}(s)
=
\begin{bmatrix}
\bar{\mathbf{p}}^{xy}
+
s\mathbf{u}
+
\widehat{n}(s)\mathbf{n}\\[1mm]
\bar{z}+\widehat{h}(s)
\end{bmatrix}.
\label{eq:global-parametric-curve}
\end{equation}
Unlike methods that constructed a curve from three selected
points~\cite{gaha2021}, the proposed formulation uses all observations
within a cluster. Note, that conductor fitting is performed using the unscaled, span-aligned local coordinates, denoted by $s_i, n_i, h_i$, and $z_i'=h_i$. The scaling in \ref{eq:adaptive-along-weight} is used exclusively for clustering.
The quadratic coefficients are estimated by iteratively reweighted least squares using Huber weights~\cite{Huber1964}, which reduces the influence of LiDAR noise and misclassified points.

For either the lateral or vertical model, let \(q_i\) denote the
observed dependent variable and \(\widehat{q}(s_i)\) the corresponding
model prediction. Specifically, \(q_i=n_i\) for the lateral model and
\(q_i=h_i\) for the vertical model. The residual is defined as
\begin{equation}
r_i=q_i-\widehat{q}(s_i).
\label{eq:model-residual}
\end{equation}
Each observation is assigned the Huber weight
\begin{equation}
\omega_i =
\begin{cases}
1,
&
|r_i|\leq c\widehat{\sigma},
\\[2mm]
\dfrac{c\widehat{\sigma}}{|r_i|},
&
|r_i|>c\widehat{\sigma},
\end{cases}
\label{eq:huber-weight}
\end{equation}
where \(c=1.345\). Let
\(\widetilde{r}=\operatorname{median}_{j}(r_j)\) denote the median of
the residuals over all observations \(j\) in the current cluster. The
robust residual-scale estimate is then computed as
\begin{equation}
\widehat{\sigma}
=
1.4826\,
\operatorname{median}_{i}
\left(
\left|
r_i-\widetilde{r}
\right|
\right).
\label{eq:mad-scale}
\end{equation}
The coefficients are updated by solving the corresponding weighted
least-squares problem. Iterations continue until the weights converge
or a maximum of \(T=10\) iterations is reached. For clusters containing at most 10 points, ordinary least squares is used because the Median Absolute Deviation (MAD)-based scale estimate could become unstable for very small samples.

\subsection{Iterative Fragment Merging and Conductor Acceptance}
\label{sec:iterative-merging}

One of the primary contributions of this work is the way we iteratively merge the powerline segments to construct a more complete representation of each physical conductor. Our approach proceeds as follows:
First, clusters are sorted by descending length, so that the longest and thus most geometrically reliable segments are processed first. Each unassigned cluster is then used in turn as a seed \(\mathcal{M}_k=\{k\}\), and the 3D parabola model of Eqs.~\eqref{eq:lateral-model} and~\eqref{eq:vertical-model} is fitted to \(\mathcal{M}_k\) using Huber-weighted regression. The compatibility of every cluster \(k'\) with this model is measured by a Huber root-mean-square error computed over all its points:
\begin{equation}
\mathrm{RMSE}_{k'}
=
\sqrt{\frac{1}{|k'|}\sum_{\mathbf{p}_i\in k'}\rho_{\delta_H}(r_i)},
\label{eq:huber-rmse}
\end{equation}
where the residual
\begin{equation}
r_i^2
=
\frac{\left(n_i-\widehat{n}(s_i)\right)^2}{1+\beta_1^2}
+
\left(h_i-\widehat{h}(s_i)\right)^2
\label{eq:merge-residual}
\end{equation}
combines the horizontal distance perpendicular to the fitted line and the vertical deviation, and
\begin{equation}
\rho_{\delta_H}(r)
=
\begin{cases}
r^2, & r\leq\delta_H,\\
2\delta_H r-\delta_H^2, & r>\delta_H,
\end{cases}
\label{eq:huber-loss}
\end{equation}
is the Huber loss with \(\delta_H=0.1\)~m, so that isolated outlying points contribute linearly rather than quadratically. A seed whose own RMSE exceeds \(\tau_{\mathrm{RMSE}}=0.30\)~m is considered incoherent and is discarded. Otherwise, the unassigned cluster with the lowest RMSE is merged into \(\mathcal{M}_k\) if its RMSE does not exceed \(\tau_{\mathrm{RMSE}}\). The model is then re-fitted on the merged point set and compared against the remaining unmerged clusters; this process repeats until no additional merge occurs. Consequently, the best-aligned fragments are merged first, and each merge refines the model used to evaluate the remaining fragments. This best-first growth reduces the risk of prematurely merging adjacent conductors. After the iterative merging procedure, each candidate conductor is evaluated according to its along-span coverage: 

\begin{equation} r_{\mathrm{span}}(C_k) = \frac{ \max_{\mathbf{p}_i \in C_k} s_i - \min_{\mathbf{p}_i \in C_k} s_i }{L}, \label{eq:span_coverage} 
\end{equation} 

where $L$ denotes the complete span length. Let $\tau_{\mathrm{span}}$ denote the minimum fraction of the complete span length that a candidate cluster must cover to be accepted as a reconstructed conductor. A candidate is accepted when 

\begin{equation} r_{\mathrm{span}}(C_k) > \tau_{\mathrm{span}}. 
\label{eq:conductor_acceptance} 
\end{equation} 

We set $\tau_{\mathrm{span}} = 0.55$. Candidates that do not satisfy this criterion are forwarded to the rescue stage.

\subsection{Rescue of Rejected Fragments}
\label{sec:rescue}
Rejected fragments are not discarded immediately since pairs of
short fragments could jointly support a valid conductor model. Let $\mathcal{R} = \left\{ F_1,\ldots,F_{N_R} \right\}$ denote the set of rejected fragments containing at least five points, having a horizontal extent of at least \(\Delta s\), and whose Huber RMSE (Eq.~\eqref{eq:huber-rmse}) with respect to their own model does not exceed \(\tau_{\mathrm{RMSE}}\).
For each unordered pair \((F_i,F_j)\), where \(i<j\), the two fragments must not overlap by more than 5\% along the span. A robust span-aligned parametric quadratic model \(\widehat{\mathcal{C}}_{ij}\) is then fitted to the combined point set \(F_i\cup F_j\). The pair is scored using the worst per-fragment Huber RMSE with respect to \(\widehat{\mathcal{C}}_{ij}\), denoted 

\begin{equation}
\eta_{ij}
=
\max
\left(
\mathrm{RMSE}_{F_i},
\mathrm{RMSE}_{F_j}
\right).
\label{eq:rescue-score}
\end{equation}
Pairs satisfying
\(\eta_{ij}\leq\tau_{\mathrm{RMSE}}\) are merged in ascending order of \(\eta_{ij}\): the best pair is merged, the pairs involving either fragment are discarded, and the merged fragment is re-scored against the remaining fragments. Each resulting candidate, merged or not, is accepted when its span coverage satisfies
\(r_{\mathrm{span}}>\tau_{\mathrm{rescue}}\), with \(\tau_{\mathrm{rescue}}=0.40\).
The rescue stage required \(O(N_R^2)\) pairwise model evaluations. 
%In the large-scale aerial evaluation, this stage increases recall by 0.2 and the F$_1$-score by 0.1, while decreasing precision by 0.1.
Algorithm~\ref{algo:reconstruction} summarizes the complete
reconstruction pipeline.

\begin{algorithm*}[p!htb]
\caption{Power-Line Reconstruction from Mobile and Aerial LiDAR Data}
\label{algo:reconstruction}

\KwIn{Semantically classified span files containing candidate wire
points}
\KwOut{Set of reconstructed conductor models}
\addtocounter{algocf}{-1}
Set
\(\epsilon_{\mathrm{HDBSCAN}}=0.3\)~m,
\(\rho_s=20\),
\(w_n=2.0\), \(w_z=1.0\),
\(w_{s,\min}=0.05\), \(w_{s,\max}=1.0\),
\(k_{\min}=3\), \(m_{\min}=3\), \(m_{\max}=15\),
\(\delta_{H}=0.1\)~m,
\(\tau_{\mathrm{RMSE}}=0.30\)~m,
\(\Delta s=0.1\)~m,
\(\tau_{\mathrm{span}}=0.55\), and
\(\tau_{\mathrm{rescue}}=0.40\)\;

\ForEach{classified span file}{
   \BlankLine
    \tcp{Stage 1: Scaled clustering and geometric filtering}
    Compute the span-aligned frame; Extract candidate wire points and compute span length \(L\)\;
    \((\mathbf{u},\mathbf{n})\) using PCA on the horizontal
    coordinates ; Project each point into local coordinates
    \((s_i,n_i,h_i)\)\;

    Compute
    \(w_s\leftarrow
    \operatorname{clip}
    \left(
    \rho_s\epsilon_{\mathrm{HDBSCAN}}/L,
    w_{s,\min},
    w_{s,\max}
    \right)\) ; Transform each point as
    \(\widetilde{\mathbf{p}}_i
    \leftarrow
    [w_ss_i,w_nn_i,w_zh_i]^{\top}\)\;

    Apply HDBSCAN with \(\epsilon_{\mathrm{HDBSCAN}}=0.3\) to
    \(\{\widetilde{\mathbf{p}}_i\}\) ; Remove  short clusters or orientation inconsistent\;

    \BlankLine
    \tcp{Stage 2: Robust parametric conductor modeling}
    Fit the lateral and vertical models in
    Eqs.~\eqref{eq:lateral-model} and~\eqref{eq:vertical-model}
    using Huber-weighted regression\;

    \BlankLine
    \tcp{Stage 3: Iterative fragment merging and acceptance}

  \ForEach{unassigned cluster $k$ (longest first)}{
      $\mathcal{M}_k \leftarrow \{k\}$\;
      \Repeat{no cluster is added to $\mathcal{M}_k$}{
      Fit parabola via Huber on $\mathcal{M}_k$: $\hat{n}(s) = \beta_1s + \beta_0$ \&  $\hat{h}(s) = \gamma_2s^2+\gamma_1s+\gamma_0$\;
      \ForEach{cluster $k'$ (seed $k$ and every unassigned cluster)}{
     Compute the Huber RMSE
     $\mathrm{RMSE}_{k'} = \sqrt{\tfrac{1}{|k'|}\sum_{p\in k'} \rho_{\delta_H}(r_p)}$, with
     $r_p^2 = \frac{(n_p-\hat{n}(s_p))^2}{1+\beta_1^2} + (h_p-\hat{h}(s_p))^2$ and
     $\rho_{\delta_H}(r) = r^2$ if $r\leq\delta_H$, else $2\delta_H r-\delta_H^2$\;
      }
      \If{$\mathcal{M}_k=\{k\}$ and $\mathrm{RMSE}_{k} > \tau_{\mathrm{RMSE}}$}{
     Discard seed $k$ (incoherent) and continue with the next seed\;
      }
      $k^{*} \leftarrow \arg\min_{k'\,\mathrm{unassigned}} \mathrm{RMSE}_{k'}$\;
      \If{$\mathrm{RMSE}_{k^{*}} \leq \tau_{\mathrm{RMSE}}$}{
     $\mathcal{M}_k \leftarrow \mathcal{M}_k \cup \{k^{*}\}$ and relabel $k^{*}$ to $k$\;
      }
      }
      Sample the conductor model of $\mathcal{M}_k$ along the span at spacing $\leq \Delta s$ to obtain the merged candidate $C_k$\;
  }

    Initialize the rejected set \(\mathcal{R}\leftarrow\varnothing\)\;

    \ForEach{merged candidate \(C_k\)}{
        Compute \(r_{\mathrm{span}}(C_k)\)\;

        \eIf{
        \(r_{\mathrm{span}}(C_k)>\tau_{\mathrm{span}}\)
        }{
            Accept and save the reconstructed conductor model\;
        }{
            Add \(C_k\) to \(\mathcal{R}\)\;
        }
    }

    \BlankLine
    \tcp{Stage 4: Rescue of rejected fragments}
    Retain fragments in \(\mathcal{R}\) having an Huber RMSE to their own model of at most \(\tau_{\mathrm{RMSE}}\)\;

    \ForEach{unordered pair \((F_i,F_j)\), \(i<j\)}{
        \If{\(F_i\) and \(F_j\) overlap by at most 5\% along the span}{
        Fit
        \(\widehat{\mathcal{C}}_{ij}\) to \(F_i\cup F_j\)\;
        Compute the rescue score
        \(\eta_{ij}=\max\left(\mathrm{RMSE}_{F_i},\mathrm{RMSE}_{F_j}\right)\)
        with respect to \(\widehat{\mathcal{C}}_{ij}\)\;
        \If{\(\eta_{ij}\leq\tau_{\mathrm{RMSE}}\)}{
            Store \((F_i,F_j,\eta_{ij})\) as a valid rescue pair\;
        }
        }
    }

    \While{valid rescue pairs remain}{
        Merge the pair with the lowest \(\eta_{ij}\) into \(F_i\cup F_j\)
        with model \(\widehat{\mathcal{C}}_{ij}\)\;

        Discard the pairs involving \(F_i\) or \(F_j\) and re-score the
        merged fragment against the remaining fragments\;
    }

    \ForEach{rescued candidate \(C_k\), merged or not}{
        \If{
        \(r_{\mathrm{span}}(C_k)>\tau_{\mathrm{rescue}}\)
        }{
            Accept and save the reconstructed conductor model\;
        }
    }
}
\end{algorithm*}

\section{Experimental Evaluation}
\label{sec:results}
We evaluate the proposed method from four complementary perspectives: 1) controlled robustness to point thinning and occlusion, 2) large-scale performance on aerial and mobile LiDAR datasets, 3) comparison with existing reconstruction pipelines, and 4) sensitivity to the main algorithmic parameters.

\subsection{Controlled Robustness Assessment}
\label{sec:controlled-evaluation}
The controlled assessment evaluates the influence of data degradation on individual power-line spans. This assessment is challenging because spans differ in geometry, length, point density, and occlusion patterns, depending on the acquisition geometry and surrounding environment. To examine these effects under controlled conditions, we select three
representative span architectures from the mobile LiDAR dataset. 
The mobile point clouds are acquired using a mapping unit equipped with two Velodyne VLP-32C LiDAR sensors, with a combined acquisition rate of two million points per second, an RTK-GPS system with a reported accuracy of \(\pm 2\)~cm, and an inertial measurement unit. The point
clouds have an average density of approximately 1,200~points/m\(^2\) and a reported positional accuracy of \(\pm 3\)~cm. The data are preprocessed and divided into individual pole-to-pole spans containing the conductors located between two consecutive poles. Further details regarding span extraction are provided in~\cite{Massicotte2026I2MTC,GRETSI2025,Massicotte2024}.
The selected architectures comprise: 1) a three-phase span,
2) a single-phase span, and 3) a horizontal-to-vertical span in which three horizontally arranged phases transition into a vertical
three-phase configuration. The third architecture represents a
complex case because the relative horizontal and vertical positions of the conductors vary along the span.

We evaluate each architecture under three degradation scenarios: uniform point thinning, multiple distributed gaps, and a single contiguous gap. The complete observations of each span define the 100\% retention condition. For each architecture, we generate 20 evaluation files, corresponding to retained-point ratios ranging from 100\% to 5\% in decrements of 5 percentage points. Each file represents a single retention level; no repeated realizations are generated at the same level. For each evaluation file, reconstructed conductors are evaluated against individual reference conductors within the same span. Let $P = \{\mathbf{p}_k\}_{k=1}^{N_P}$ denote the points sampled along a predicted conductor curve, and let $G$ denote the sampled reference geometry of a single reference conductor. The three-dimensional Euclidean distance from each predicted sample to its nearest reference point is 

\begin{equation} d_k = \min_{\mathbf{g} \in G} \left\| \mathbf{p}_k - \mathbf{g} \right\|_2. 
\label{eq:prediction_reference_distance} 
\end{equation} 

The geometric match ratio is defined as 

\begin{equation} \rho(P,G) = \frac{1}{N_P} \sum_{k=1}^{N_P} \mathbb{I}\!\left( d_k \leq d_{\mathrm{match}} \right), \qquad d_{\mathrm{match}} = 0.20\,\mathrm{m}, 
\label{eq:geometric_match_ratio} 
\end{equation} 

where $\mathbb{I}(\cdot)$ equals one when its argument is true and zero otherwise. A prediction--reference pair is eligible for matching when
\begin{equation} \rho(P,G) \geq \tau_{\mathrm{match}}, \qquad \tau_{\mathrm{match}} = 0.70. 
\label{eq:conductor_matching_criterion} 
\end{equation} 

Thus, at least 70\% of the predicted samples must lie within 0.20\,m of the corresponding reference conductor. One-to-one matching is enforced within each realization: each prediction can be assigned to at most one reference conductor, and each reference conductor can be assigned to at most one prediction. Assigned prediction--reference pairs are counted as true positives (TP). Unmatched predictions are counted as false positives (FP), and unmatched reference conductors are counted as false negatives (FN).

\subsubsection{Robustness to Uniform Point Thinning}
\label{sec:uniform-thinning}
Uniform point thinning evaluates the effect of progressively reducing the number of available observations without explicitly introducing contiguous occlusion intervals. Points are randomly retained according to the specified retention ratio. The original conductors contain
approximately 25--30 points/m. 
Fig.~\ref{fig:controlled-results} reports reconstruction F$_1$ score as a function of the retained-point ratio. All three span architectures achieve 100\% performance at retention ratios of 5\% or greater. At a 5\% retention ratio, the effective linear density is approximately 1.25--1.5 points/m. When the retention ratio falls below 5\%, the remaining observations no longer provide sufficient geometric support
for reliable conductor reconstruction.

\subsubsection{Robustness to Multiple Distributed Gaps}
\label{sec:distributed-gaps}
\begin{figure}[pos=!htb]
    \centering
\includegraphics[width=\columnwidth]{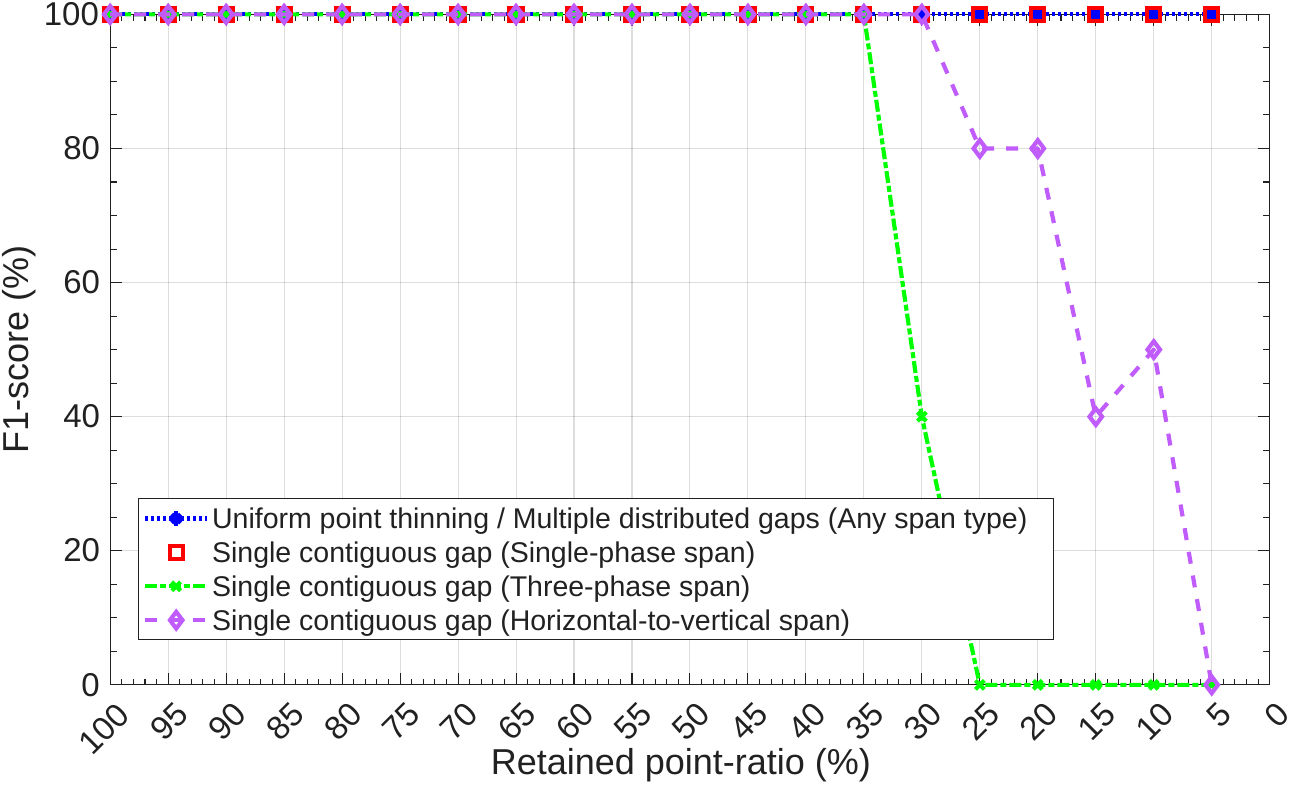}
    \caption{Reconstruction F$_1$-score for three conductor architectures under controlled degradation: uniform point thinning, multiple distributed gaps, and a single contiguous gap.}
    \label{fig:controlled-results}
\end{figure}

\begin{figure}[pos=!htb]
\centering
\includegraphics[width=0.85\columnwidth]{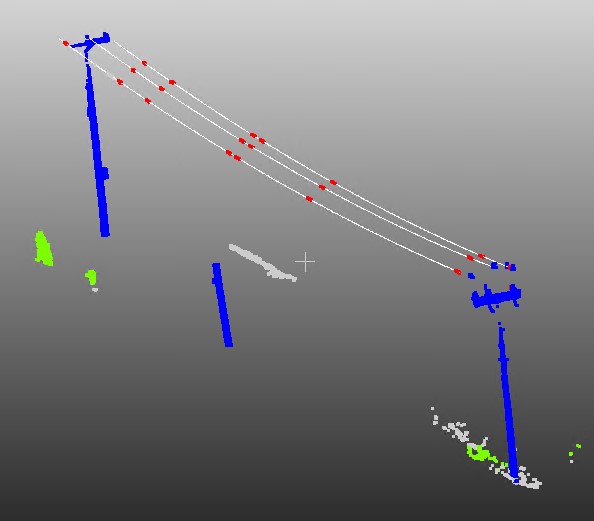}
\caption{Reconstruction of a three-phase span using 5\% of the original
conductor observations (red: retained LiDAR observations, white: reconstructed conductor).}
\label{fig:five-percent}
\end{figure}

The second scenario introduces multiple gaps at randomly selected
locations along each conductor to reproduce fragmented observations
caused by vegetation occlusion. Unlike uniform point thinning, which distributes the retained observations across the complete span, this scenario removes contiguous groups of neighboring points while preserving the specified overall retention ratio.
Fig.~\ref{fig:controlled-results} shows that the proposed method
achieves F$_1$ score of 100\% for all three architectures at retention ratios of 5\% or greater. Below this threshold, the remaining fragments become too short and too widely separated to support reliable model fitting and fragment association. Fig.~\ref{fig:five-percent} illustrates a successful reconstruction of a three-phase span using only 5\% of the original conductor observations.

\subsubsection{Robustness to a Single Contiguous Gap}
\label{sec:single-gap}
\begin{figure}[pos=!htb]
\centering
\includegraphics[width=\columnwidth]{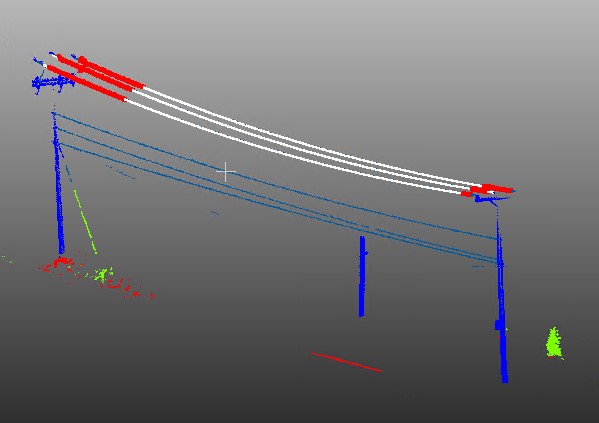}
\caption{Example of a single contiguous occlusion covering 75\% of the conductor span (red: retained LiDAR observations, white: reconstructed conductor).}
\label{fig:large-occlusion}
\end{figure}

The third scenario evaluates reconstruction F$_1$ score in the presence of a single contiguous gap centered within a 58-m power-line span. This experiment examines whether the proposed method can associate conductor fragments separated by a long occluded interval. The gap is progressively enlarged toward both ends of the span, thus reducing the retained-point ratio and the length of the observed conductor fragments. As shown in Fig.~\ref{fig:controlled-results}, all three architectures
achieve F$_1$ score of 100\% at retention ratios of 35\% or greater. At a retention ratio of 30\%, our method no longer reconstructs the three-phase architecture successfully, although the other tested architectures remain detectable. At retention ratios below 10\%, the
remaining fragments are too short and separated to support reliable reconstruction for any of the three architectures. Fig.~\ref{fig:large-occlusion} illustrates a case in which 75\% of the conductor span is occluded.

\subsection{Large-Scale Evaluation}
\label{sec:large-scale-evaluation}
The large-scale assessment evaluates the proposed method on
heterogeneous aerial and mobile LiDAR datasets. Since independent
conductor-level reference annotations are available only for the aerial dataset, the two modalities use different evaluation protocols. The aerial evaluation measures object-level detection and geometric reconstruction performance, whereas the mobile evaluation measures point-to-model geometric consistency and reconstruction coverage.

\subsubsection{Aerial LiDAR Evaluation}
\label{sec:aerial-evaluation}
The aerial dataset contains 10,870 pole-to-pole spans covering 382~km of distribution networks. It includes single-phase and
three-phase systems, mixed horizontal and vertical configurations, and both medium-voltage and low-voltage infrastructure. Individual span
lengths range from 4 to 160~m. The point clouds are acquired using a RIEGL VQ-1460 laser scanner
operating at a pulse repetition rate of up to 4.4~MHz. The point density ranges from 1 to 30~points/m\(^2\), and the reported vertical RMSE over
flat surfaces is 0.05~m. The LiDAR service provider supplies one 3D reference geometry for each physical conductor within each pole-to-pole span. The dataset contains 23,372 reference conductors. The reference geometries and predicted conductor models are expressed in the same coordinate reference system. Since the complete details of the provider's proprietary annotation procedure are unavailable, we refer to these geometries as provider-supplied reference annotations rather than error-free ground truth.

Prediction-to-reference matching is performed indepen-
dently within each span as defined in Subsection \ref{sec:controlled-evaluation}. For each successfully matched pair, the distance from each reference point to the predicted curve is computed. All resulting point-to-curve distances from successfully matched pairs are pooled before computing the reported geometric RMSE. Table~\ref{tab:aerial-performance} summarizes the large-scale aerial results. The proposed method reconstructs 22,592 conductors that matche to provider-supplied reference conductors. Our method achieves an F$_1$-score of 96.76\%, and a geometric RMSE of 0.127~m.
\begin{table*}[pos=!htb]
\centering
\caption{Large-scale performance on the aerial LiDAR dataset.}
\label{tab:aerial-performance}
\small
\setlength{\tabcolsep}{5pt}
\renewcommand{\arraystretch}{1.1}
\begin{tabular}{ccccccccc}
\hline
\textbf{Reference} &
\textbf{Predicted} &
\textbf{TP} &
\textbf{FP} &
\textbf{FN} &
\textbf{Precision (\%)} &
\textbf{Recall (\%)} &
\textbf{F$_1$ (\%)} &
\textbf{RMSE (m)} \\
\hline
23,372 &
23,323 &
22,592 &
731 &
780 &
96.87 &
96.66 &
96.76 &
0.127 \\
\hline
\end{tabular}
\end{table*}

Table~\ref{tab:aerial-statistics} summarizes the distribution of the per-span F$_1$-score, processing time, and point-to-reference distance, defined as the distance between provider-supplied reference conductor points and the corresponding predicted parabola. Although the maximum point-to-reference distance reaches 3~m, a closer examination of the results revealed that this value corresponds to a single outlier located near the end of a span and is therefore not representative of the overall reconstruction accuracy. Indeed, 95\% of the points exhibit a point-to-reference distance below 0.112~m, indicating a high level of geometric agreement between the predicted and reference conductors.

\begin{table*}[pos=!htb]
\centering
\caption{Distribution of the per-span F$_1$-score, processing time and geometric error on the aerial LiDAR dataset.}
\label{tab:aerial-statistics}
\small
\setlength{\tabcolsep}{4pt}
\renewcommand{\arraystretch}{1.1}
\begin{tabular}{lccccc}
\hline
\textbf{Metric} & \textbf{Mean} & \textbf{Std} & \textbf{Median} & \textbf{P95} & \textbf{Max} \\
\hline
F$_1$ per span (\%)               & 96.31 & 15.82 & 100.00 & 100.00 & 100.00 \\
Processing time per span (s)      & 0.153 & 0.342 & 0.100  & 0.350  & 1.47 \\
Point-to-reference distance (m)   & 0.056 & 0.091 & 0.040  & 0.112  & 3.00 \\
RMSE per conductor (m)            & 0.095 & 0.072 & 0.073  & 0.234  & 1.04 \\
\hline
\end{tabular}
\end{table*}

\subsubsection{Mobile LiDAR Evaluation}
\label{sec:mobile-evaluation}
The mobile dataset contains 1,620 power-line spans covering 68~km of
distribution networks. It includes single-phase and three-phase
systems, mixed horizontal and vertical layouts, and medium-voltage and low-voltage infrastructure. Individual span lengths range from 4 to 120~m. As a reminder, the 180 development spans are excluded from the 1,620 test spans. Independent conductor-level reference annotations are unavailable for this dataset. Consequently, the mobile evaluation does not measure object-level precision, recall, or F$_1$-score. Instead, it evaluates
the geometric consistency between the candidate wire points produced by the semantic-classification stage and the reconstructed conductor models using reconstruction coverage and point-to-model RMSE. For each candidate wire point
\(\mathbf{p}_i\), let $e_i
=\min_{\widehat{\mathbf{p}}\in\widehat{\mathcal{C}}}\left\|
\mathbf{p}_i-\widehat{\mathbf{p}}
\right\|_2$ denote its Euclidean distance between LiDAR conductor points and all reconstructed conductor curves \(\widehat{\mathcal{C}}\) in the corresponding span. A candidate wire point is considered geometrically covered when
\begin{equation}
e_i\leq d_{\mathrm{cov}},
\qquad
d_{\mathrm{cov}}=0.15~\mathrm{m}.
\label{eq:mobile-coverage-condition}
\end{equation}
Let \(N_{\mathrm{wire}}\) denote the total number of candidate wire
points and \(N_{\mathrm{covered}}\) the number satisfying eq.~\eqref{eq:mobile-coverage-condition}. Reconstruction coverage is
defined as $\mathrm{Coverage}
=\frac{N_{\mathrm{covered}}}{N_{\mathrm{wire}}}
\times 100\%$. The point-to-model RMSE is computed over the covered candidate wire points as
\begin{equation}
\mathrm{RMSE}_{\mathrm{mobile}}
=
\sqrt{
\frac{1}{N_{\mathrm{covered}}}
\sum_{i:e_i\leq d_{\mathrm{cov}}}
e_i^2
}.
\label{eq:mobile-rmse}
\end{equation}
The proposed method achieves a reconstruction coverage of 95.27\% and a point-to-model RMSE of 0.045~m. Thus, most points classified as wires are geometrically supported by a reconstructed conductor model within a maximum distance of 0.15~m.

\subsection{Comparative Benchmarking}
\label{sec:comparative-evaluation}
We compare the proposed method with two existing reconstruction
pipelines using the controlled dataset described in
Subsection~\ref{sec:controlled-evaluation}. The comparison includes the same three conductor architectures and three degradation scenarios as defined for Fig. \ref{fig:controlled-results}. %, i.e., uniform point thinning, multiple distributed gaps, and a single contiguous gap.

%\subsubsection{Compared Methods}
%\label{sec:compared-methods}
The first baseline is a publicly available MATLAB implementation derived from the method of Shi et al.~\cite{Shi2020}. It detects and models power lines using geometric feature extraction, curve fitting, and plane-rotation operations. We refer to this implementation as EPL. 
The second quantitative baseline is our reimplementation of the method of Cao et al.~\cite{2025IJSTA}, hereafter referred to as Cao-FEC.
Because the original implementation is unavailable, the results reflect our implementation based on the methodological details reported in~\cite{2025IJSTA}. The original method was developed for transmission corridors and reported an F$_1$-score of 99.79\% on its evaluation datasets. It relies on Fast Euclidean Clustering (FEC) with a distance
threshold \(D_{\mathrm{FEC}}\) to separate individual conductors. 
When directly applied to the distribution-network benchmark, the
Cao-FEC implementation achieves an F$_1$-score of 47.7\%. This
performance difference is explained by the conflicting distance conditions within the evaluated distribution spans. To avoid merging neighboring conductors, the clustering threshold must satisfy
\begin{equation}
D_{\mathrm{FEC}}
<
d_{\mathrm{inter}}^{\min}
\approx
1.0~\mathrm{m},
\label{eq:fec-upper-bound}
\end{equation}
where \(d_{\mathrm{inter}}^{\min}\) denotes the minimum observed
separation between neighboring conductors. However, associating
fragments belonging to the same conductor requires $D_{\mathrm{FEC}} > d_{\mathrm{gap}}^{\max} \approx
2.4~\mathrm{m}$, where \(d_{\mathrm{gap}}^{\max}\) denotes the maximum observed gap between fragments of the same conductor. Since these conditions cannot
be satisfied simultaneously in the original coordinate system, direct
Euclidean clustering either fragments individual conductors or merges neighboring phases.

We also evaluate a modified Cao-FEC configuration using the
span-aligned prescaling described in Subsection~\ref{sec:scaled-clustering}. With
\(w_s=0.5\), \(w_n=2.0\), \(w_z=1.0\), and an RMSE threshold of \(T_{\mathrm{RMSE}}=0.5\)~m, the modified implementation achieves an F$_1$-score of 89.7\% on the controlled benchmark without changing the core FEC procedure. This result shows that span-aligned prescaling partially mitigates the conflicting distance constraints, although its performance remains below that of the proposed method.

%\subsubsection{Comparative Results}
%\label{sec:comparative-results}

\begin{figure}[pos=!htb]
\centering
\includegraphics[width=\columnwidth]
{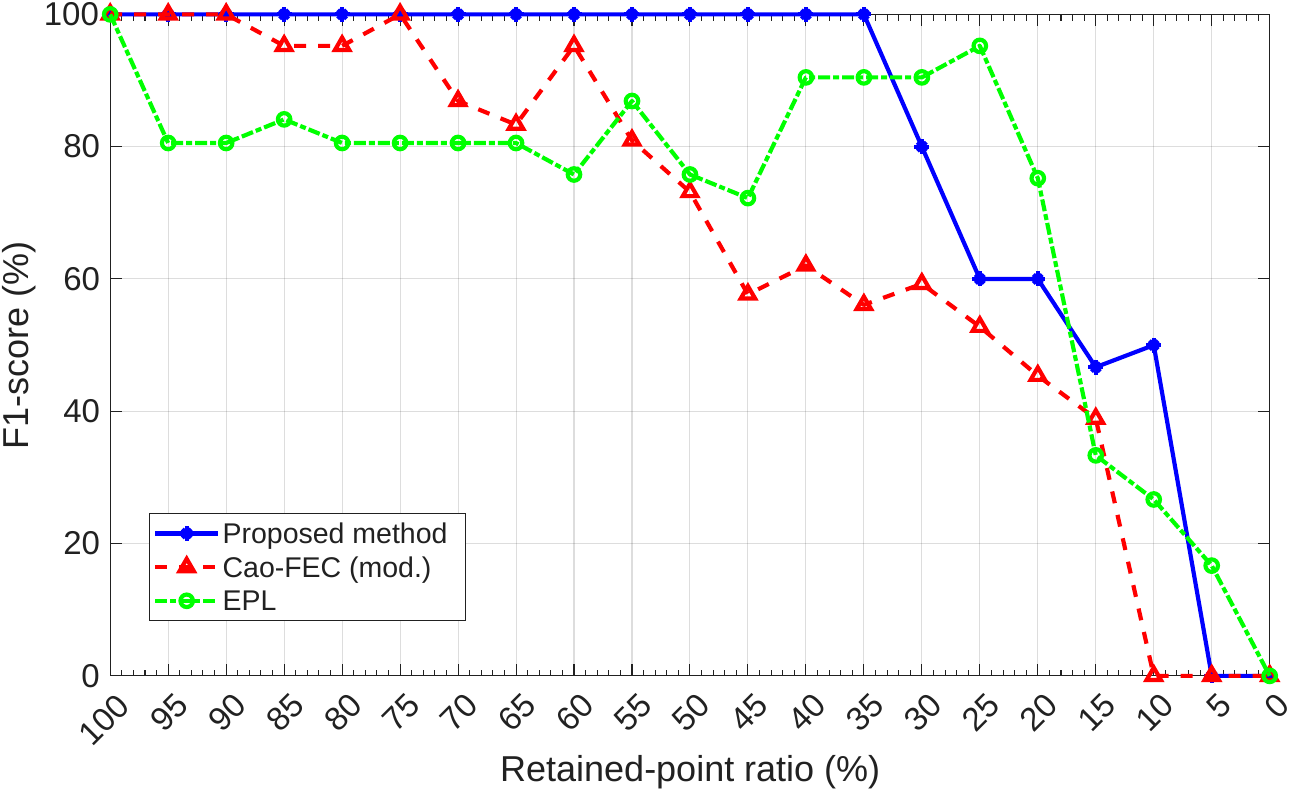}
\caption{Comparison of the proposed method, EPL, and Cao-FEC (mod.) under a
single contiguous gap for mobile LiDAR.}
\label{fig:comparison-main-gap}
\end{figure}

\begin{figure}[pos=!htb]
\centering
\includegraphics[width=\columnwidth]
{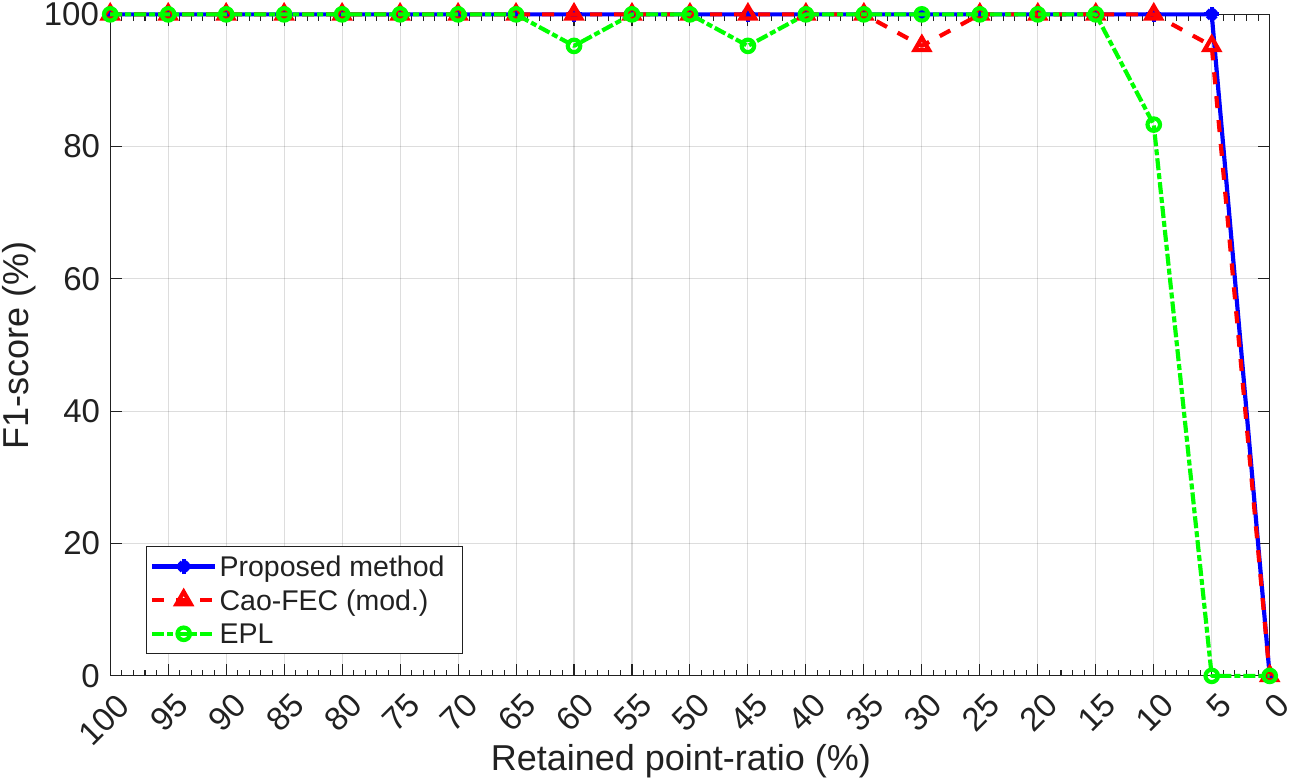}
\caption{Comparison of the proposed method, EPL, and Cao-FEC (mod.) under
uniform point thinning for mobile LiDAR.}
\label{fig:comparison-density}
\end{figure}

\begin{figure}[pos=!htb]
\centering
\includegraphics[width=\columnwidth]
{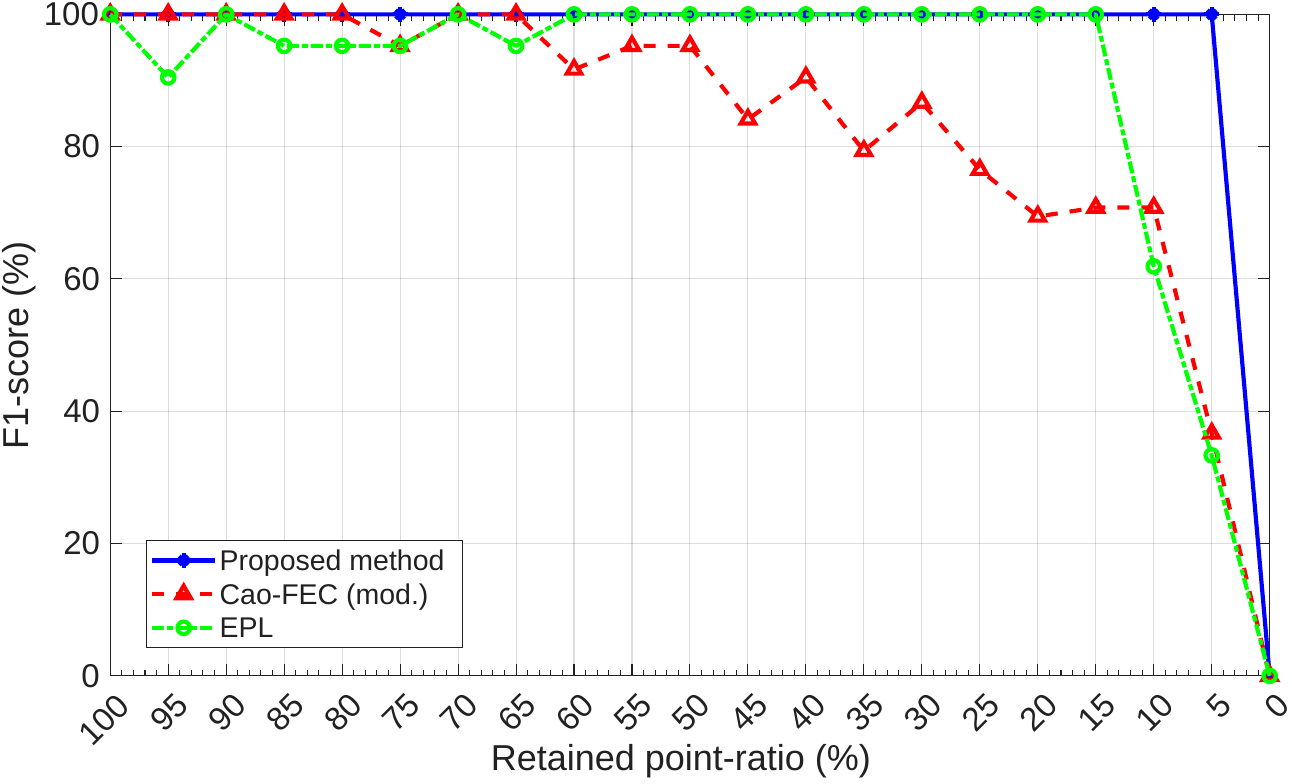}
\caption{Comparison of the proposed method, EPL, and Cao-FEC (mod.) under
multiple distributed gaps for mobile LiDAR.}
\label{fig:comparison-small-gaps}
\end{figure}

Figs.~\ref{fig:comparison-main-gap}--\ref{fig:comparison-small-gaps} compare the proposed method, EPL, and the modified Cao-FEC (Cao-FEC mod.) implementation under the three degradation scenarios. The results are aggregated by scenario over the three conductor architectures and the corresponding degraded realizations.

Under the single contiguous gap scenario, the proposed method maintains an F$_1$-score of 100\% for retained-point ratios between 35\% and 100\%, as shown in Fig.~\ref{fig:comparison-main-gap}. The performance of
Cao-FEC (mod.) decreases sharply when the retained-point ratio reaches 60\%, indicating sensitivity to increasing separation between fragments. EPL recovers some conductors under severe degradation and achieves
F$_1$-scores between 90\% and 95\% for certain configurations at
retention ratios between 25\% and 35\%. However, persistent FP predictions reduce its overall performance across the complete benchmark. Thus, EPL can outperform the proposed method in a
small number of severely degraded configurations, but the proposed
method provides more consistent performance across architectures and retention ratios. 
Figs.~\ref{fig:comparison-density} and
\ref{fig:comparison-small-gaps} present the results for uniform point thinning and multiple distributed gaps, respectively. The proposed method achieves the highest overall F$_1$-scores across both scenarios. This advantage results primarily from its progressive fragment-merging
procedure, which associates short but geometrically compatible
fragments while limiting premature merging between adjacent phases.

%These comparative results indicate that the proposed method is more stable than the evaluated baselines under sparse and fragmented observations. Nevertheless, they should be interpreted in the context of the controlled benchmark. In particular, Cao-FEC (mod.) is our reimplementation of a method originally designed for transmission corridors, whereas the present benchmark contains closely spaced distribution conductors.

\begin{table*}[pos=!htb]
\centering
\caption{Selected parameters of the proposed reconstruction pipeline.}
\label{tab:parameters}
\small
\setlength{\tabcolsep}{5pt}
\renewcommand{\arraystretch}{1.1}
\begin{tabular}{p{2.4cm} p{8.3cm} p{2.1cm} p{1.5cm}}
\hline
\textbf{Parameter} &
\textbf{Description} &
\textbf{Selected value} &
\textbf{Type} \\
\hline
\(\epsilon_{\mathrm{HDBSCAN}}\) &
HDBSCAN cluster-selection epsilon in the scaled coordinate space &
0.3 &
Fixed \\

\(k_{\min}\) &
HDBSCAN minimum number of samples &
3 &
Fixed \\

\(m_{cs}\) &
HDBSCAN minimum cluster size depending on points density &
\([3,15]\) &
Adaptive \\

\(w_n\) &
Scale weight applied to the cross-span coordinate &
2 &
Fixed \\

\(w_z\) &
Scale weight applied to the vertical coordinate &
1 &
Fixed \\

\(w_{s,\min}\) &
Lower bound on the adaptive along-span scale weight &
0.05 &
Fixed \\

\(w_{s,\max}\) &
Upper bound on the adaptive along-span scale weight &
1 &
Fixed \\

\(\rho_s\) &
Scaling coefficient used to derive the adaptive along-span weight &
20 &
Fixed \\

\(w_s\) &
Adaptive along-span scale weight, clipped to
\([w_{s,\min},w_{s,\max}]\) &
\([0.05,1]\) &
Adaptive \\

\(d_{\mathrm{seg,min}}\) &
Minimum accepted along-span extent of an initial cluster &
0.1~m &
Fixed \\

\(N_{\mathrm{wire,min}}\) &
Minimum number of candidate wire points required to process a span &
20 &
Fixed \\

\(\delta_{H}\) &
Huber threshold of the RMSE used for merging and rescue &
0.1~m &
Fixed \\

\(\tau_{\mathrm{RMSE}}\) &
Maximum Huber RMSE of a fragment with respect to a fitted conductor model for merging and rescue &
0.3~m &
Fixed \\

\(\Delta s\) &
Maximum sampling spacing of the conductor model and minimum extent of a rescue fragment &
0.1~m &
Fixed \\

\(\tau_{\mathrm{span}}\) &
Minimum fraction of the complete span length required for acceptance &
0.55 &
Fixed \\

\(\tau_{\mathrm{rescue}}\) &
Minimum fraction of the complete span length required for accepting a rescued candidate &
0.40 &
Fixed \\
\hline
\end{tabular}
\end{table*}

\subsection{Sensitivity Analysis}
\label{sec:sensitivity}

A subset of 180 synthetically generated spans, comprising 420 conductors, was used exclusively for parameter selection and sensitivity analysis. The corresponding ground-truth annotations were generated with the synthetic data. These spans are excluded from the 1,620 mobile test set spans and are also excluded from the large-scale aerial evaluation set containing 10,870 spans. The resulting parameter configuration is subsequently applied to the large-scale dataset without further modification. 
We vary one parameter at a time while holding all remaining parameters at their selected values. Derived parameters are recomputed according to their defining equations. Performance is measured against the provider-supplied reference annotations using the matching protocol defined in Subsection~\ref{sec:aerial-evaluation}. 

\subsubsection{Sensitivity to HDBSCAN Cluster-Selection Epsilon}
\label{sec:sensitivity-epsilon}

Table~\ref{tab:epsilon-sensitivity} reports performance as
\(\epsilon_{\mathrm{HDBSCAN}}\) varies from 0.1 to 2.0. Across this
range, the F$_1$-score varies from 96.5\% to 97.5\%, indicating
low sensitivity to this parameter, and no monotonic trend is observed. In fact, TP varies only between 399 and 403 and FP between 4 and 9. The selected value,
\(\epsilon_{\mathrm{HDBSCAN}}=0.3\), lies within 0.4 points of the highest
F$_1$-score, obtained at \(\epsilon_{\mathrm{HDBSCAN}}=0.9\). This gain does not transfer to the airborne dataset, where \(\epsilon_{\mathrm{HDBSCAN}}=0.9\) reduces the F$_1$-score from 96.76\% to 96.56\% and increases the reconstruction of RMSE from 0.127 to 0.17~m; \(\epsilon_{\mathrm{HDBSCAN}}=0.3\) is therefore retained for both datasets. This robustness is partly explained by the subsequent fragment-merging stage, which reassembles conductors that HDBSCAN splits into short fragments, and by the span-coverage criterion, which rejects most incoherent clusters.
Since $w_s$ depends on $\epsilon_{\mathrm{HDBSCAN}}$, this experiment evaluates their coupled influence on reconstruction performance. In particular, the scaled along-wire extent \(w_s L=\rho_s\,\epsilon_{\mathrm{HDBSCAN}}\) grows with \(\epsilon_{\mathrm{HDBSCAN}}\). Further experiments with independently controlled scaling and clustering parameters are required to distinguish their individual effects.

\begin{table}[pos=!htb]
\centering
\caption{Sensitivity to HDBSCAN cluster-selection epsilon for mobile LiDAR.}
\label{tab:epsilon-sensitivity}
\small
\setlength{\tabcolsep}{3pt}
\renewcommand{\arraystretch}{1.1}
\begin{tabular}{ccccccc}
\hline
\(\boldsymbol{\epsilon_{\mathrm{HDBSCAN}}}\) &
\textbf{TP} &
\textbf{FP} &
\textbf{FN} &
\textbf{Prec.} &
\textbf{Rec.} &
\textbf{F$_1$} \\
\hline
0.1 & 402 & 4  & 18 & 99.0 & 95.7 & 97.3 \\
0.2 & 400 & 7  & 20 & 98.3 & 95.2 & 96.7 \\
0.3 & 402 & 6  & 18 & 98.5 & 95.7 & 97.1 \\
0.4 & 402 & 6  & 18 & 98.5 & 95.7 & 97.1 \\
0.5 & 400 & 9  & 20 & 97.8 & 95.2 & 96.5 \\
0.7 & 399 & 7  & 21 & 98.3 & 95.0 & 96.6 \\
\textbf{0.9} & \textbf{403} & \textbf{4} & \textbf{17}
& \textbf{99.0} & \textbf{96.0} & \textbf{97.5} \\
1.0 & 401 & 5  & 19 & 98.8 & 95.5 & 97.1 \\
1.5 & 402 & 6  & 18 & 98.5 & 95.7 & 97.1 \\
2.0 & 401 & 7  & 19 & 98.3 & 95.5 & 96.9 \\
\hline
\end{tabular}
\end{table}

\subsubsection{Sensitivity to the Huber RMSE Merging Threshold}
\label{sec:sensitivity-rmse}

During iterative merging, a fragment is appended to a seed curve only if the Huber RMSE ($\delta_H = 0.1$~m) of the joint fit satisfies $\mathrm{RMSE}_H \leq \tau_{\mathrm{RMSE}}$. 
This same threshold is applied to discard incoherent seeds and validate rescue candidates. 
Table~\ref{tab:rmse-sensitivity} reports the resulting performance as $\tau_{\mathrm{RMSE}}$ varies from 0.05 to 1.0~m. 

The reconstruction performance remains stable within the interval $0.10 \leq \tau_{\mathrm{RMSE}} \leq 0.35$~m, yielding an $F_1$-score between 96.6\% and 97.3\%. 
The optimal result is achieved at $\tau_{\mathrm{RMSE}} = 0.20$~m, and the adopted baseline of $\tau_{\mathrm{RMSE}} = 0.30$~m lies well within this stable plateau. 
Strictly lower thresholds over-reject valid fragment associations, causing conductors to remain fragmented into multiple distinct curves and consequently increasing false positives (FPs). 
Conversely, excessively large thresholds erroneously permit the merging of fragments from physically distinct conductors. 
As a result, both precision and recall drop sharply for thresholds exceeding 0.4~m.

\begin{table}[pos=!htb]
\centering
\caption{Sensitivity to the Huber RMSE merging threshold for mobile LiDAR.}
\label{tab:rmse-sensitivity}
\small
\setlength{\tabcolsep}{3pt}
\renewcommand{\arraystretch}{1.1}
\begin{tabular}{ccccccc}
\hline
\(\boldsymbol{\tau_{\mathrm{RMSE}}}\) &
\textbf{TP} &
\textbf{FP} &
\textbf{FN} &
\textbf{Prec.} &
\textbf{Rec.} &
\textbf{F$_1$} \\
\hline
0.05 & 401 & 24  & 19  & 94.4 & 95.5 & 94.9 \\
0.10 & 401 & 9   & 19  & 97.8 & 95.5 & 96.6 \\
0.15 & 402 & 8   & 18  & 98.0 & 95.7 & 96.9 \\
\textbf{0.20} & \textbf{403} & \textbf{5} & \textbf{17}
& \textbf{98.8} & \textbf{96.0} & \textbf{97.3} \\
0.25 & 401 & 6   & 19  & 98.5 & 95.5 & 97.0 \\
0.30 & 402 & 6   & 18  & 98.5 & 95.7 & 97.1 \\
0.35 & 401 & 8   & 19  & 98.0 & 95.5 & 96.7 \\
0.40 & 387 & 15  & 33  & 96.3 & 92.1 & 94.2 \\
0.50 & 134 & 119 & 286 & 53.0 & 31.9 & 39.8 \\
0.70 & 59  & 115 & 361 & 33.9 & 14.0 & 19.9 \\
1.00 & 59  & 115 & 361 & 33.9 & 14.0 & 19.9 \\
\hline
\end{tabular}
\end{table}

\begin{figure}[pos=!htb]
\centering
\includegraphics[width=\columnwidth]{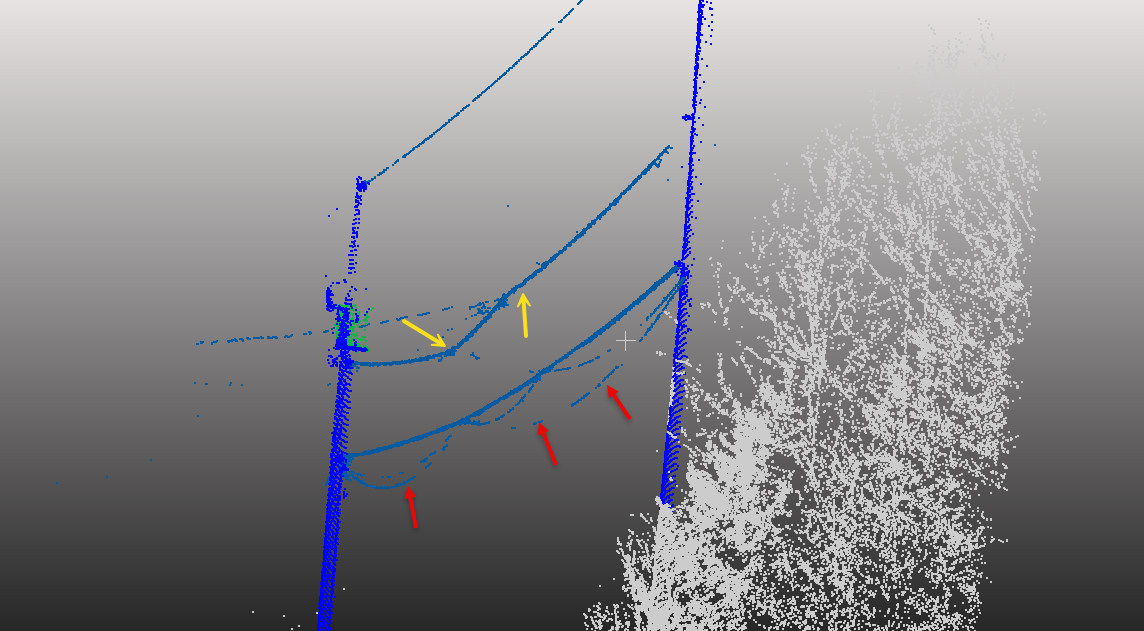}
\caption{Challenging configurations involving intertwined conductors
and nonlinear conductor trajectories.}
\label{fig:challenging-configurations}
\end{figure}

\subsubsection{Influence of the Curve Model and Fitting Estimator}
\label{sec:sensitivity-fitting}

In the following experiments, a single component of the pipeline is modified at a time, while all other parameters are kept at the values listed in Table~\ref{tab:parameters}. Table~\ref{tab:fitting-sensitivity} compares the default Huber-weighted polynomial fitting with two alternatives: (i) the same line and parabola estimated by ordinary least squares (OLS) without re-weighting, and (ii) a catenary model as defined in equation \ref{eq:catenary}: 
\begin{equation}
\label{eq:catenary}
z(t)=z_0+\frac{\cosh\!\left(k(t-t_m)+\varphi\right)-\cosh(\varphi)}{k}
\end{equation}

estimated with the same Huber IRLS scheme. The fitting model is used consistently in fragment merging, span validation and rescue.

Omission of the Huber re-weighting induces the most detrimental effect: the $F_1$-score decreases from 97.1\% to 94.8\%, accompanied by a simultaneous increase in both false positives (FPs) and false negatives (FNs). 
Without robust re-weighting, stray points bias the fitted curve, artificially inflating the RMSE of valid fragments. 
Consequently, these fragments fail to merge with their corresponding conductors, resulting in omitted conductors and residual fragments misclassified as FPs. 

The catenary model yields comparable but slightly inferior performance relative to the parabolic model ($F_1$-score of 96.5\% versus 97.1\%). 
Crucially, it exhibits a 3.6-fold increase in computational runtime. 
This overhead stems from the fact that each IRLS iteration requires a non-linear least-squares solution, whereas the parabolic model remains linear in its parameters. 
Given that the two curves differ by only a few millimeters under typical spans and sags, the parabola serves as a highly sufficient approximation. 

\begin{table}[pos=!htb]
\centering
\caption{Influence of the curve model and fitting estimator for mobile LiDAR.}
\label{tab:fitting-sensitivity}
\small
\setlength{\tabcolsep}{3pt}
\renewcommand{\arraystretch}{1.1}
\begin{tabular}{lccccccc}
\hline
\textbf{Fitting} &
\textbf{TP} &
\textbf{FP} &
\textbf{FN} &
\textbf{Prec.} &
\textbf{Rec.} &
\textbf{F$_1$}  \\
\hline
Polynomial (OLS)       & 394 & 17 & 26 & 95.9 & 93.8 & 94.8  \\
Catenary (Huber IRLS)  & 400 & 9  & 20 & 97.8 & 95.2 & 96.5 \\
\textbf{Polynomial (Huber IRLS)} & \textbf{402} & \textbf{6} & \textbf{18}
& \textbf{98.5} & \textbf{95.7} & \textbf{97.1} &  \\
\hline
\end{tabular}
\end{table}

\subsubsection{Standard versus Huber RMSE}
\label{sec:sensitivity-huber-rmse}

To evaluate the merging criterion, Eq.~\eqref{eq:huber-rmse} replaces the standard squared residual with the Huber loss ($\delta_H = 0.1$~m). 
Consequently, residuals exceeding $\delta_H$ contribute only linearly to the total error. 
Table~\ref{tab:huber-rmse-sensitivity} compares this approach with the standard RMSE across various thresholds ($\tau_{\mathrm{RMSE}}$), while keeping all other pipeline stages identical.

At the selected threshold ($\tau_{\mathrm{RMSE}} = 0.30$~m), both criteria yield nearly identical performance: the standard RMSE achieves an $F_1$-score of 97.3\%, whereas the Huber RMSE achieves 97.1\%, representing a difference of a single conductor. 
However, because the Huber loss grows more slowly than the squared loss, it produces lower values for identical residuals. 
For a given $\tau_{\mathrm{RMSE}}$, the Huber criterion is therefore more permissive. 
This behavior explains why its performance degrades more rapidly at larger thresholds; at $\tau_{\mathrm{RMSE}} = 0.40$~m, the $F_1$-score drops to 94.2\% compared to 97.0\% for the standard RMSE, as the algorithm begins erroneously merging fragments from neighboring conductors.

On this synthetic dataset, which contains few outliers, the bounded influence of the Huber loss provides no measurable advantage. 
Nevertheless, its primary utility lies in processing real-world data, where it prevents a small number of stray points from causing the rejection of an otherwise coherent line fragment.

\begin{table}[pos=!htb]
\centering
\caption{Standard versus Huber RMSE merging criterion for mobile LiDAR. The adopted configuration is the Huber RMSE with \(\tau_{\mathrm{RMSE}}=0.30\)~m.}
\label{tab:huber-rmse-sensitivity}
\small
\setlength{\tabcolsep}{3pt}
\renewcommand{\arraystretch}{1.1}
\begin{tabular}{c|cccc|cccc}
\hline
 & \multicolumn{4}{c|}{\textbf{Standard RMSE}} & \multicolumn{4}{c}{\textbf{Huber RMSE}} \\
\(\boldsymbol{\tau_{\mathrm{RMSE}}}\) &
\textbf{TP} & \textbf{FP} & \textbf{FN} & \textbf{F$_1$} &
\textbf{TP} & \textbf{FP} & \textbf{FN} & \textbf{F$_1$} \\
\hline
0.10 & 401 & 9 & 19 & 96.6 & 401 & 9  & 19 & 96.6 \\
0.15 & 401 & 8 & 19 & 96.7 & 402 & 8  & 18 & 96.9 \\
0.20 & 403 & 5 & 17 & 97.3 & 403 & 5  & 17 & 97.3 \\
0.25 & 403 & 5 & 17 & 97.3 & 401 & 6  & 19 & 97.0 \\
0.30 & 403 & 5 & 17 & 97.3 & \textbf{402} & \textbf{6} & \textbf{18} & \textbf{97.1} \\
0.40 & 401 & 6 & 19 & 97.0 & 387 & 15 & 33 & 94.2 \\
\hline
\end{tabular}
\end{table}

\subsubsection{Effect of Span-Aligned Prescaling}
\label{sec:prescaling-ablation}
Distribution network conductors in LiDAR data can extend over more than 50~m while being separated laterally by only 0.7--1.0~m. Applying HDBSCAN directly to the original coordinates treats the along-span and cross-span coordinates equally. Consequently, large longitudinal gaps can dominate the distance metric, while the smaller separation between neighboring conductors receives insufficient emphasis. 
In the unscaled baseline, all coordinate weights are set to one, i.e., \(w_s=w_n=w_z=1\), which reduces the prescaling to a rotation of the point cloud. In the proposed configuration, \(w_s\in[0.05,1.0]\) is computed adaptively, \(w_n=2.0\) and \(w_z=1.0\); all other parameters are unchanged.
Table~\ref{tab:prescaling-ablation} compares both configurations. On the mobile LiDAR dataset, prescaling has a small effect: it recovers 2 additional TPs at the cost of 1
additional FP, increasing the F$_1$-score by 0.1. The synthetic conductors are densely and regularly sampled, so that most fragments are already separated in the original coordinates. On the aerial dataset, where the point density is lower and longitudinal gaps are larger, the prescaling increases the F$_1$-score and reduces the reconstruction RMSE.
%This improvement results from reducing along-span distances between compatible fragments while increasing the relative lateral and vertical separation between neighboring conductors.

\begin{table}[pos=!htb]
\centering
\caption{Detection performance with and without span-aligned
prescaling for mobile and aerial LiDAR. Precision, recall, and F$_1$-score are reported as percentages.}
\label{tab:prescaling-ablation}
\small
\setlength{\tabcolsep}{4pt}
\renewcommand{\arraystretch}{1.1}
\begin{tabular}{l|l|cccccc}
\hline
\textbf{Dataset} &
\textbf{Configuration} &
\textbf{Precision(\%)} &
\textbf{Recall(\%)} &
\textbf{F$_1$(\%)} \\
\hline
Mobile & Prescaling   & 98.5 & \textbf{95.7} & \textbf{97.1} \\
       & No prescaling  & \textbf{98.8} & 95.2 & 97.0 \\
\hline
Aerial & Prescaling   & \textbf{99.2} & \textbf{99.4} & \textbf{99.3} \\
       & No prescaling & 98.6 & 95.9 & 97.2 \\
\hline
\end{tabular}
\end{table}

\subsubsection{Overall Parameter Sensitivity}
\label{sec:overall-sensitivity}

Table~\ref{tab:sensitivity-summary} summarizes the one-at-a-time parameter sensitivity analysis, with all other parameters fixed at their selected values (F$_1$=97.1\%). The results show that \(\tau_{\mathrm{RMSE}}\) and \(\delta_H\) have the strongest influence on performance, followed by \(\tau_{\mathrm{span}}\). The parameters \(\tau_{\mathrm{RMSE}}\) and \(\delta_H\) directly affect fragment merging. A small \(\tau_{\mathrm{RMSE}}\) rejects valid associations between separated fragments, whereas a large value can incorrectly merge neighboring conductors (Section~\ref{sec:sensitivity-rmse}). 

Similarly, decreasing \(\delta_H\) reduces the RMSE of candidate associations and therefore has an effect comparable to increasing \(\tau_{\mathrm{RMSE}}\). At \(\delta_H=0.02\)~m, neighboring phases are merged in all multi-conductor spans, whereas values \(\delta_H\geq0.2\)~m produce identical results (F$_1$=97.3\%). The parameter \(\tau_{\mathrm{span}}\) is stable between 0.2 and 0.8, but values approaching 1 reduce recall by requiring conductors to cover nearly the entire span. In contrast, the clustering-related parameters show limited sensitivity, with F$_1$ variations of at most one percentage point. The parameter \(w_{s,\max}\) has no effect because \(w_s\) never exceeds 0.1 for the span lengths considered in this dataset. A few alternative settings slightly improve the mobile LiDAR results. Testing these settings on the airborne dataset did not yield consistent improvements. The selected parameter values were therefore retained as a balanced configuration for both mobile and airborne LiDAR data.

\begin{table}[pos=!htb]
\centering
\caption{Summary of the one-at-a-time parameter sensitivity analysis for mobile LiDAR.}
\label{tab:sensitivity-summary}
\small
\setlength{\tabcolsep}{6pt}
\renewcommand{\arraystretch}{1.1}
\begin{tabular}{lcccc}
\hline
\textbf{Param.} &
\textbf{Selec. Val.} &
\textbf{Test. range} &
\textbf{Min. F$_1$} &
\textbf{Max. F$_1$} \\
\hline
\(\tau_{\mathrm{RMSE}}\)         & 0.30 & \([0.05,1]\)   & 19.9 \% & 97.3 \% \\
\(\delta_H\)                     & 0.10 & \([0.02,0.5]\) & 19.9 \% & 97.3 \% \\
\(\tau_{\mathrm{span}}\)         & 0.55 & \([0.2,1]\)    & 90.9 \% & 97.1 \% \\
\(w_{s,\max}\)                   & 1    & \([0.1,2]\)    & 97.1 \% & 97.1 \% \\
\(w_{s,\min}\)                   & 0.05 & \([0.01,0.5]\) & 97.1 \% & 97.5 \% \\
\(\epsilon_{\mathrm{HDBSCAN}}\)  & 0.3  & \([0.1,2]\)    & 96.5 \% & 97.5 \% \\
\(\rho_s\)                       & 20   & \([5,200]\)    & 96.9 \% & 97.5 \% \\
\(k_{\min}\)                     & 3    & \([2,30]\)     & 96.5 \% & 97.3 \% \\
\hline
\end{tabular}
\end{table}

\section{Conclusion and Future Work}
\label{sec:conclusion}

This paper presents a robust method for reconstructing power-line geometry from sparse and occluded mobile and aerial LiDAR point clouds. Following semantic classification, the proposed pipeline applies the same reconstruction implementation and parameter configuration to both modalities. It combines span-aligned prescaling, HDBSCAN clustering, Huber-weighted polynomial fitting, iterative fragment merging, and a rescue stage for rejected fragments.

The controlled experiments demonstrate robustness to uniform point thinning, distributed gaps, and large contiguous occlusions. Under these conditions, the proposed method provides more consistent performance than the evaluated EPL and Cao-FEC pipelines. The sensitivity analysis identifies the Huber RMSE threshold as the most influential parameter, while showing relatively stable performance across the tested HDBSCAN epsilon values. 

On 10,870 aerial spans containing 23,372 reference conductors, our method achieved a precision of 96.87\%, a recall of 96.66\%, an F$_1$-score of 96.76\%, and a geometric RMSE of 0.12~m. On 1,620 mobile spans, it achieved a reconstruction coverage of 95.27\% and a point-to-model RMSE of 0.045~m. The main limitations are the absence of independent conductor-level annotations for the mobile dataset. However, the controlled robustness experiments and the aerial evaluation use independent conductor references and therefore partially mitigate this limitation. Another limitation is the reliance on generated semantic labels for aerial and mobile LiDAR, and the limited capability of the quadratic model to represent nonlinear, crossing, or intertwined conductors such as in Fig.~\ref{fig:challenging-configurations}. 

Future work will investigate orientation-aware hierarchical clustering and more flexible geometric models for these configurations. Moreover, it will strengthen the integration of conductor reconstruction with vegetation-encroachment assessment and utility decision-support tools
\cite{Massicotte2026I2MTC}. To support reproducibility, we make the source code, documentation, and evaluation data publicly available at \url{https://github.com/varen-project}.

\section*{CRediT authorship contribution statement}

\textbf{Mohamed Gaha}: Conceptualization, Methodology, Investigation, Formal analysis, Writing -- original draft. \textbf{Damos Ayobo Abongo}: Validation, Formal analysis, Visualization, Writing -- review \& editing. \textbf{Philippe Massicotte}: Validation, Writing -- review \& editing. \textbf{Eya Ben Kacem}: Validation, Formal analysis, Writing -- review \& editing. \textbf{Wael Jaafar}: Supervision, Writing -- review \& editing. \textbf{Rami Langar}: Supervision, Writing -- review \& editing. \textbf{Christian Buteau}: Supervision, Project administration.

\section*{Acknowledgements}
This work is supported by the NSERC Alliance-Mitacs Accelerate research program CRSNG ALLRP 607469-25.

\bibliographystyle{cas-model2-names}
\bibliography{BibLio}

\end{document}